\documentclass{article}

\usepackage{microtype}
\usepackage{graphicx}
\usepackage{subfigure}
\usepackage{booktabs} 

\usepackage{hyperref}

\usepackage[accepted]{icml2024}

\usepackage{amsmath}
\usepackage{amssymb}
\usepackage{mathtools}
\usepackage{amsthm}

\usepackage[capitalize,noabbrev]{cleveref}

\theoremstyle{plain}
\newtheorem{theorem}{Theorem}[section]

\theoremstyle{definition}
\newtheorem{definition}[theorem]{Definition}
\newtheorem{assumption}[theorem]{Assumption}
\theoremstyle{remark}

\usepackage{colortbl}
\usepackage[textsize=tiny]{todonotes}
\usepackage{multirow}
\usepackage{graphicx}
\usepackage{xcolor}
\usepackage{tcolorbox}
\DeclareMathOperator*{\argmax}{argmax}
\newcommand*{\fullref}[1]{\hyperref[{#1}]{\ref*{#1}}}
\usepackage[normalem]{ulem}
\useunder{\uline}{\ul}{}
\usepackage{anyfontsize}
\usepackage{xspace}
\newcommand{\method}{WCA\xspace}

\usepackage[inline]{enumitem}
\usepackage{booktabs,tabularx}
\usepackage[export]{adjustbox}
\usepackage{wrapfig}
\usepackage{pifont}
\newcommand{\xmark}{\ding{55}}%
\newcommand{\cmark}{\ding{51}}%
\usepackage{bm}
\definecolor{custompink}{rgb}{1, 0.2, 0.6}

\icmltitlerunning{Visual-Text Cross Alignment: Refining the Similarity Score in Vision-Language Models}

\begin{document}

\twocolumn[
\icmltitle{Visual-Text Cross Alignment: \texorpdfstring{\\Refining the Similarity Score in Vision-Language Models}{}}



\icmlsetsymbol{equal}{*}

\begin{icmlauthorlist}
\icmlauthor{Jinhao Li}{unimelb}
\icmlauthor{Haopeng Li}{unimelb}
\icmlauthor{Sarah M. Erfani}{unimelb}
\icmlauthor{Lei Feng}{sutd}
\icmlauthor{James Bailey}{unimelb}
\icmlauthor{Feng Liu}{unimelb}
\end{icmlauthorlist}

\icmlaffiliation{unimelb}{School of Computing and Information Systems, University of Melbourne, Australia.}
\icmlaffiliation{sutd}{Information Systems Technology and Design Pillar, Singapore University of Technology and Design, Singapore}

\icmlcorrespondingauthor{Feng Liu}{fengliu.ml@gmail.com}

\icmlkeywords{visual-text cross alignment, zero-shot performance, vision-language models, visual prompting, CLIP, zero-shot classification, visual-text similarity, large language models, fine-grained descriptions, visual-language alignment, similarity matrix, image-text alignment, cross-modal learning, zero-shot learning, visual prompting techniques}

\vskip 0.3in
]


\printAffiliationsAndNotice{}  

\begin{abstract}
It has recently been discovered that using a pre-trained \emph{vision-language model} (VLM), e.g., CLIP, to align a whole query image with several finer text descriptions generated by a large language model can significantly enhance zero-shot performance. 
However, in this paper, we empirically find that the finer descriptions tend to align more effectively with \emph{local areas of the query image} rather than the whole image, and then we theoretically validate this finding.
Thus, we present a method called \emph{weighted visual-text cross alignment} (\method).
This method begins with a \emph{localized visual prompting} technique, designed to identify local visual areas within the query image.
The local visual areas are then \emph{cross-aligned} with the finer descriptions by creating a similarity matrix using the pre-trained VLM. 
To determine how well a query image aligns with each category, we develop a score function based on the weighted similarities in this matrix.
Extensive experiments demonstrate that our method significantly improves zero-shot performance across various datasets, achieving results that are even comparable to few-shot learning methods.
The code is available at \href{https://github.com/tmlr-group/WCA}{\textcolor{custompink}{github.com/tmlr-group/WCA}}.
\end{abstract}

\section{Introduction}\label{sec:intro}
Following the significant advancements of large-scale pre-training in natural language processing \cite{devlin2018bert, radford2018improving, radford2019language, brown2020language}, the CLIP model \cite{radford2021learning} scales up its pre-training data through \emph{aligning} images and the corresponding natural language captions in the shared latent space, which achieves remarkable performance in zero-shot classification. Despite its achievements, CLIP's performance exhibits notable sensitivity to the prompts used during the inference stage \cite{radford2021learning, zhou2022learning}. For example, \citet{zhou2022learning} has highlighted that changing the prompt from ``a photo of [CLASS]" to ``a photo of a [CLASS]"  can lead to a performance boost of 6\%. Crafting effective prompts is crucial but it requires significant time, effort, and domain-specific knowledge \cite{zhou2022learning}, making it challenging to deploy such models in practical applications.

\begin{figure}[t]
\begin{center}
\centerline{\includegraphics[width=0.98\columnwidth]{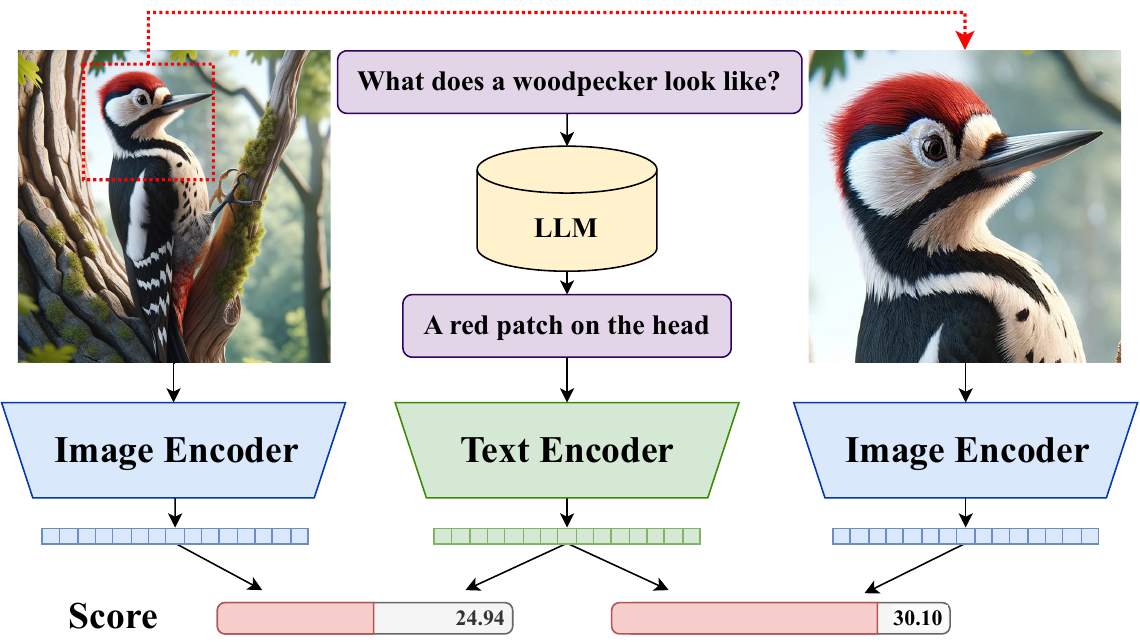}}
\vspace{-0.5em}
\caption{\textbf{Aligning an entire image with a detailed text description results in lower scaled cosine similarity}, as shown on the left. Aligning the description with a specific image part, such as the detailed red patch (on the right), increases the score.}
\label{fg:align}
\end{center}
\vspace{-2.7em}
\end{figure}

To address the above issue, a promising solution is to use \emph{large language models} (LLMs) to generate several finer text descriptions of each category \cite{menon2022visual, pratt2023does}. This strategy helps reduce the manual effort in creating prompts and, more importantly, does not necessitate additional turning, thereby more easily preserving models' generalization abilities. The ability to generalize is a crucial issue in prompt-learning methods \cite{li2022ordinalclip, wang2022learning, wu2023pi, tanwisuthZZHZ23} as these methods tend to overfit training data. LLM-based visual-text alignment emphasizes \emph{global matching}, namely, text descriptions are aligned with the whole image.

\begin{figure*}[t]
\begin{center}
\centerline{\includegraphics[width=\textwidth]{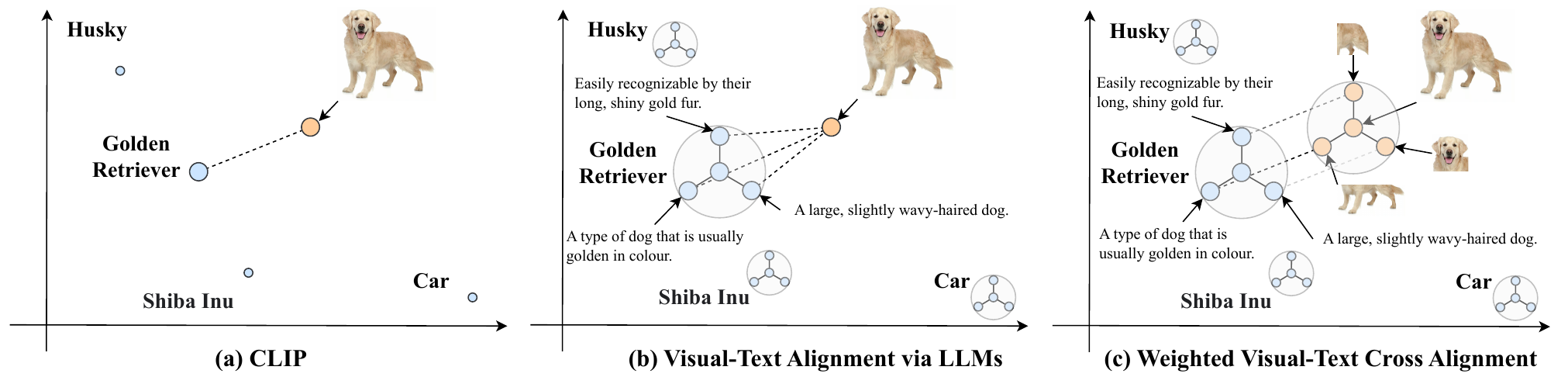}}
\caption{\textbf{We show different zero-shot visual-text alignment methods}: (a) CLIP, (b) Visual-Text Alignment via LLMs \cite{menon2022visual, pratt2023does}, and (c) Weighted Visual-Text Cross Alignment (ours). Unlike (a) and (b), (c) utilizes a localized visual prompting technique to enhance alignment by ensuring that detailed descriptions match precisely with specific areas of the visual content.}
\label{motivation}
\end{center}
\vskip -0.2in
\end{figure*}

However, in this paper, we find: 
\begin{tcolorbox}[left=1pt,right=1pt,top=0pt,bottom=0pt,boxrule=0.5mm,before skip=0.5em, after skip=0em,]
\emph{Finer-grained text descriptions may align more accurately with the specific area of an image but not necessarily with the image as a whole.}
\end{tcolorbox}

This is because finer descriptions often contain detailed and finer-grained visual concepts, such as ``a woodpecker has a straight and pointed bill", may not align precisely with an entire image, as demonstrated in \figureautorefname~\ref{fg:align}. 
In expanding this, it becomes clear that the complexity of images often contains a myriad of details that a description might capture only in part. These finer elements, while crucial, might lead to a misalignment when the objective is to correlate the entire image with its description. Additionally, we provide a theoretical analysis in Section~\ref{sec:motivation} to gain a deeper understanding of the issue. This discrepancy suggests a need for more nuanced approaches in visual-text alignment, where the focus is not just on the whole query image (i.e., the global alignment) but also on recognizing and aligning different areas within the query image. 

To this end, we propose a method called \emph{\textbf{w}eighted visual-text \textbf{c}ross \textbf{a}lignment} (\method). This method concentrates on local and regional visual areas within the query image that can align better with the fine-grained text descriptions of each category. This can be achieved through \emph{localized visual prompting}, where the image is prompted to focus on localized visual elements, such as via cropping. These visual elements are then \emph{cross-aligned} with their corresponding detailed text descriptions, leading to a similarity matrix. Furthermore, we introduce a score function based on the \emph{weighted similarities} in the matrix and use the score to see how well a query image is aligned with each category. 

A key feature of \method is its consideration of the varying importance of localized visual elements and text descriptions within the similarity matrix, ensuring that each similarity value in this matrix contributes differently to the overall similarity aggregation. The importance of a specific localized visual element is quantified by its cosine similarity with the query image, where a high score suggests that the element captures the main semantic content of the query image. Similarly, the relevance of a text description can be estimated by its similarity to the category label it corresponds to. 
Therefore, our method achieves accurate visual-text alignment scores efficiently without additional models or training, making it a highly efficient approach. 
We demonstrate the distinction between \method and the current methods of visual-text alignment in \figureautorefname~\ref{motivation}. 
Our empirical results show that \method significantly enhances zero-shot performance across various datasets, even comparable with few-shot methods.

To summarize, our main contributions are outlined as follows:
\begin{enumerate*}[label=(\roman*)]
  \item We have identified and conducted a theoretical analysis of the issue where aligning an entire image with finer text descriptions results in suboptimal performance.
  \item We introduce a method, \method, that performs \emph{weighted cross alignment} between the finer descriptions and local visual areas using localized visual prompting.
  \item Our extensive experiments validate our theoretical hypothesis and demonstrate the efficacy of our method, significantly surpassing the state-of-the-art methods without the need for extra models or data.
  \item We offer insights into the key factors that contribute to the effectiveness of our method.
\end{enumerate*}

\section{Related Work}

\textbf{Vision-language models.}
Vision-language models (VLMs) pre-trained on large-scale data have shown efficacy in enhancing representation learning capabilities \cite{cho2021unifying, kim2021vilt, xue2021probing, li2021align, wang2021simvlm,li2023progressive}. CLIP \cite{radford2021learning} underwent training on a corpus of 400 million paired images and texts, exhibiting robust transferable ability and exceptional zero-shot performance. In a similar vein, the introduction of ALIGN \cite{jia2021scaling} demonstrates that despite being pre-trained on datasets containing image-text pairs with considerable noise, the scale of the training corpus can compensate for this noise and is capable of learning superior representations. Subsequent works, including FLAVA \cite{singh2022flava}, Florence \cite{yuan2021florence}, BLIP \cite{li2022blip}, and so on, have continued to advance this paradigm, contributing further to the field.
 
\textbf{Textual prompting in vision-language models.}
Despite CLIP exhibiting superior zero-shot capabilities, the effectiveness of its application in downstream tasks is significantly influenced by the choice of prompts, as noted by \citet{radford2021learning} and \citet{zhou2022learning}. \citet{zhou2022learning} highlight that selecting the optimal prompt is complex and time-intensive, often requiring prompt tuning. To address this, \citet{menon2022visual} and \citet{pratt2023does} leverage the knowledge embedded in LLMs, such as GPT-3 \cite{brown2020language} for the automatic generation of class-specific descriptions. These descriptions, particularly focusing on the discriminating features of image categories, are then aligned with the query image. This method has been shown to be effective as it enriches the textual representation by incorporating LLMs. \citet{roth2023waffling} examine this phenomenon and introduce the WaffleCLIP framework, which replaces LLM-generated descriptions with random character and word descriptions, eliminating the need to query LLMs and offering a cost-effective alternative. However, our work diverges from these approaches as they do not engage in visual prompting techniques.

\textbf{Visual prompting in vision-language models.}
In contrast to text-based prompting in VLMs, visual prompting aims to process the visual input accordingly. \citet{yao2021cpt} color image regions and utilize a captioning model to identify objects based on color predictions. \citet{bahng2022exploring} experiment with learning the image perturbation, keeping the model parameters unchanged. These approaches, along with the studies \cite{jia2022visual, tu2023visual} require at least a few samples from downstream tasks. The RedCircle introduced by \citet{shtedritski2023does} suggests that highlighting an object with a red circle can direct the model's attention to that region, but it requires manual annotation. \citet{yang2023fine} propose FGVP, which uses an extra model SAM \cite{kirillov2023segment}, to identify objects first and then employ \emph{Blur Reverse Masks} to enhance the semantic localization capability of areas around the objects, reducing the need for manual annotation but adding complexity. Our method differs from these as we do not need downstream data, manual annotation, or additional models. 

\textbf{Test time prompt tuning in vision-language models. } 
While fine-tuning prompts can adapt pre-trained VLMs to specific downstream tasks, this approach requires labeled training data, which can be costly and unavailable for zero-shot tasks. Test-time prompt tuning (TPT), as introduced by \citet{shu2022test}, addresses this issue by learning adaptive prompts for individual test samples through the generation of multiple randomly augmented views. The goal is to optimize text prompts in an unsupervised manner. However, naive augmentation methods may lead to overly simplistic variations in test data. To address this, \citet{feng2023diverse} proposed DiffTPT, which uses diffusion models to augment test samples with richer visual appearance variations. 
In contrast, our method, while also applied during testing, does not involve the same tuning processes as TPT and DiffTPT. Instead, it leverages the strengths of pre-trained VLMs in a different manner, potentially offering a more efficient approach. Our method avoids the need for extensive data augmentation and fine-tuning procedures, which are typically required by TPT and DiffTPT to enhance their performance. By directly utilizing the inherent capabilities of pre-trained VLMs, our approach simplifies the alignment process.

\begin{figure*}[!t]
\begin{center}
\centerline{\includegraphics[width=\textwidth]{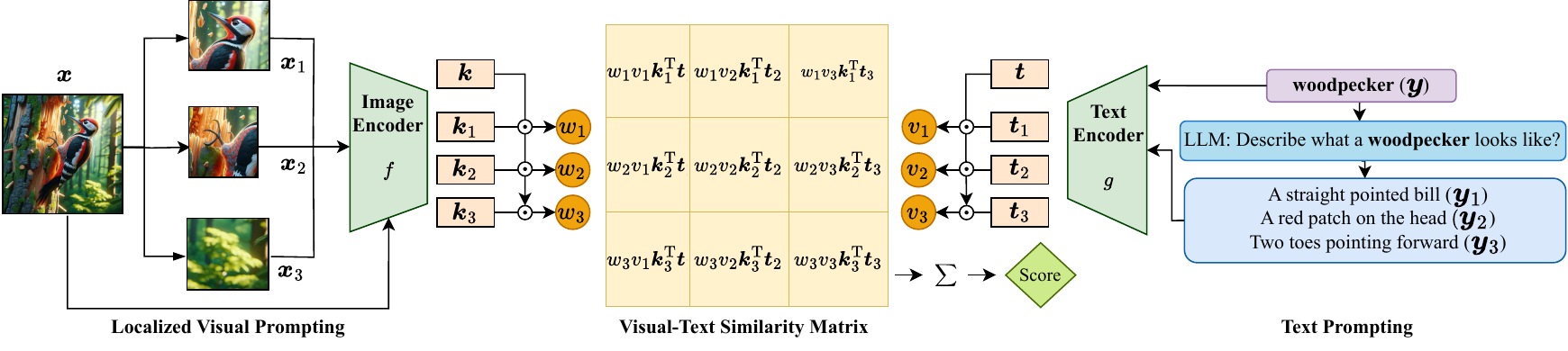}}
\caption{\textbf{Overview of weighted visual-text cross alignment (\method)}. The process begins with \emph{localized visual prompting}, where the input image $\bm x$ is divided into localized patches, such as $\{\bm x_1, \bm x_2, \bm x_3\}$. These patches are encoded by an image encoder to produce visual features. The \emph{text prompting} stage utilizes a large language model to generate detailed textual descriptions $\{\bm y_1, \bm y_2, \bm y_3\}$ for a given class label $\bm y$ (e.g., ``woodpecker"). The \method calculates alignment scores between visual features and textual features, using patch weights $\{w_1, w_2, w_3\}$ and text weights $\{v_1, v_2, v_3\}$. The final score is computed by summing the visual-text similarity matrix.}
\label{fg:overview}
\end{center}
\vskip -0.3in
\end{figure*}

\section{Problem Setting and Preliminaries}
In this section, we introduce the problem setting and the preliminaries considered in this paper.

\textbf{Problem setting.} Let $\mathcal{X}$ be an image space and $\mathcal{Y}$ be a label space, where $\mathcal{Y}$ is a set of words or phrase, e.g., $\mathcal{Y}=\{\text{car},\dots,\text{bicycle}\}$. Considering a pre-trained VLM, let $f\colon \mathcal{X} \rightarrow \mathbb{R}^d$ be its image encoder and $g\colon \mathcal{Y} \rightarrow \mathbb{R}^d$ be its text encoder. These encoders are designed to transform input images and texts into a shared embedding space of dimension $d$. In this paper, $\bm x$ represents an arbitrary image from $\mathcal{X}$, and $\bm y$ denotes an arbitrary label from $\mathcal{Y}$.

The aim in \emph{zero-shot visual classification} is to label images into predefined classes based on their visual content, without updating the parameters of the pre-trained model. We will introduce two representative methodologies to address the zero-shot visual classification problem in the following.

\textbf{CLIP zero-shot transfer} \cite{radford2021learning}. The core idea is to devise a scoring function $\mathcal{X} \times \mathcal{Y} \rightarrow \mathbb{R}$ that can assess the semantic matching between the given image and a set of corresponding labels. $s(\cdot)$ is computed based on the cosine similarity of their hidden representations. The scoring function is mathematically expressed as:
\begin{equation}
    s(\bm x, \bm y |f,g) = \mathrm{cos}(f(\bm x), g(\bm y)),
\end{equation} where a higher score implies a closer semantic match between $\bm x$ and $\bm y$. Therefore, the predicted label for image $\bm x$ is the label $\bm y^*$, which has the highest cosine similarity score with $\bm x$ among all possible labels from $\mathcal{Y}$.

\textbf{Enhancing zero-shot transfer using LLMs} \cite{pratt2023does,menon2022visual}. Given a label $\bm y \in \mathcal{Y}$, an LLM model $h(\cdot)$ can be utilized to generate rich and descriptive text that encapsulates the characteristics and details of the category $\bm y$. The descriptions are as follows:
\begin{equation}
\label{y_ens}
    h(\bm y) = \{{\bm y}_j\}_{j=1}^{M},
\end{equation} where $M$ represents the total number of generated descriptions. In this case, the scoring function $s$ is calculated as the average of similarity scores between $\bm x$ and each text description $\bm y_j$. This can be mathematically represented as: 
\begin{align}
    s_{\text{LLM}}(\bm x, \bm y |f,g) = \frac{1}{M} \sum\nolimits_{j=1}^M s(\bm x, {\bm y}_j |f,g).
\end{align}

\section{Motivation from Theoretical Justification}
\label{sec:motivation}
In Section~\ref{sec:intro}, we have empirically shown that aligning a whole image with finer text descriptions might cause a lower similarity compared to aligning an area of the image with the finer description (see Figure~\ref{fg:align}). 
In this section, we gain a deeper understanding of the issue and provide a theoretical analysis regarding the issue mentioned in Figure~\ref{fg:align}.

For simplicity, we assume that the image encoder $f$ is linear functional and satisfies the condition\footnote{This means that when $\bm{x}$ represents a non-black image, its representation is generally not a zero vector, which is a relatively weak assumption.} $\bm x \neq \bm 0 \Rightarrow f(\bm x) \neq \bm 0$. We focus on cosine similarity as we investigate CLIP-like models.
Specifically, we have the following theorem (For the complete proof, please refer to Appendix \ref{proof}).
\begin{theorem}
\label{thm:bigtheorem}
Let $\bm x$ represent an image along with its corresponding ground truth label $\bm y$. $\bm x$ can be partitioned into two components $\bm x_1$ and $\bm x_2$, where $\bm x = \bm x_1 + \bm x_2$. Assume $\bm x_1$ is a discriminative region that is perfectly correlated with $\bm y$ as $\cos(f(\bm x_1), g(\bm y)) = 1$, and non-discriminative region $\bm x_2$ has an imperfect correlation to $\bm y$ denoted as $\cos(f(\bm x_2), g(\bm y)) < 1$. If $\bm x_1$ and $\bm x_2$ satisfy linear independence\footnote{It means that there is not a constant $c$ such that ${\bm x_1}=c{\bm x_2}$. Intuitively, for an image of a cat in a garden, where $\bm x_1$ represents the cat and $\bm x_2$ denotes the garden. The information about the cat (like its shape, color, etc.) is exclusive to $\bm x_1$, and the information about the garden (like plants, sky, etc.) is exclusive to $\bm x_2$. $\bm x_1$ and $\bm x_2$ are linearly independent as neither can be represented by the other. Therefore, this assumption is relatively weak.}, then we have $\cos(f(\bm x), g(\bm y)) < 1$.
\end{theorem} 
This highlights a possible limitation in the current methodology of visual-text alignment, where encoding the entire image content might lead to a less-than-ideal performance. Therefore, it becomes essential to accurately retrieve $\bm x_1$, ensuring its semantic content is perfectly correlated with $\bm y$. A simple method to tackle this issue is to choose the highest cosine similarity score, expressed as $\max(\cos(f({\bm x}_1), g(\bm y)), \cos(f({\bm x}_2), g(\bm y)))$, to determine the most accurate alignment. However, this approach might not always be feasible, particularly if $\bm x_1$ shows a perfect similarity score to an incorrectly matched $\bm y$, leading to potential errors, which is validated in \tableautorefname~\ref{tab:aggregation}. Motivated by this, we propose \method to address this issue.

\section{Visual-text Cross Alignment}
In this section, we formally introduce our proposed method \method, where the overall pipeline is shown in Figure \ref{fg:overview}. Specifically, we start by describing localized visual prompting and then discuss how to perform the weighted cross alignment. Finally, we show the overall algorithm.

\textbf{Localized visual prompting.}
As described previously, matching an entire image with finer text descriptions could result in a lower similarity score compared to aligning a specific area of the image with the finer description. To tackle this issue, we propose \emph{localized visual prompting}, which seeks to segment an image into multiple areas, each holding critical semantic content. The objective of this method is to enhance the extraction of semantic information from images by focusing on specific regions rather than the entire image.

This can be achieved through a localized visual prompting function $p(\cdot)$.
Given an image $\bm x \in \mathbb{R}^{H \times W \times 3}$, where $H$ and $W$ are its height and width respectively, the function $p(\cdot)$ can be described as follows:
\begin{equation}
\label{x_patch}
    p_{}(\bm x)=\left\{{\bm x}_i=\phi(\bm x,\gamma_i \min(W,H))\mid i=1,...,N\right\},
\end{equation} 
where $\gamma_i$ is a random variable sampled from a uniform distribution $U(\alpha, \beta)$. $\alpha$ and $\beta$ are predefined parameters that set the lower and upper bound.
The function $\phi(\cdot)$ crops the image at a random location, with the second argument specifying the size of the output. The random nature of $\gamma_i$ ensures that the cropping is varied, covering different parts of the image, thus retrieving the different semantic information from the various regions. These localized image patches are then \emph{cross-aligned} with finer text descriptions.

\begin{figure}[t]
\begin{center}
\centerline{\includegraphics[width=0.75\columnwidth]{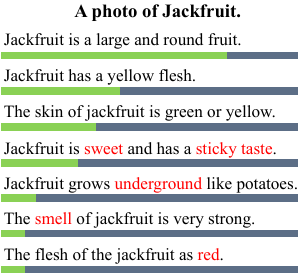}}
\caption{\textbf{Overview of text description weights for Jackfruit}. This figure illustrates various text description weights based on their relevance to the visual cue ``A photo of Jackfruit." The green lines represent the relative similarity scores, indicating how closely each description aligns with the visual cue. Longer green lines denote higher relevance, while shorter lines indicate lower relevance. Descriptions deemed irrelevant or incorrect are highlighted in red.}
\label{fg:text_weight}
\end{center}
\vskip -0.3in
\end{figure}

\textbf{Cross alignment.}
Upon obtaining the set of localized image patches $p(\bm x)$ and the set of text descriptions $h(\bm y)$, it is essential to evaluate the similarities between them, a process referred to as \emph{cross alignment}. This process results in a matrix defined as:
\begin{equation}
\label{cross-alignment}
    \begin{bmatrix}
    s({\bm x}_1, {\bm y}_1) & \cdots & s({\bm x}_1, {\bm y}_{M})\\
    \vdots & \ddots & \vdots\\
    s({\bm x}_{N}, {\bm y}_1) & \cdots & s({\bm x}_{N}, {\bm y}_{M})
    \end{bmatrix},
\end{equation} where each column's entries represent the similarity scores between the $j$-th text description and every image patch, while the entries in each row denote the scores of the $i$-th image patch and all text descriptions. A naive approach to aggregate this matrix is by averaging all scores as follows,
\begin{equation}
\label{navie_eq}
    s_{\text{AVG}}(\bm x, \bm y |f,g)=\frac{1}{NM} \sum_{i=1}^N\sum_{j=1}^M s(\bm x_i, \bm y_j |f,g).
\end{equation} 
The issue with Eq.~\eqref{navie_eq} lies in treating each image patch or description \emph{equally} in the calculation of the final similarity score. However, \citet{menon2022visual} demonstrated that with such scoring LLMs can result in suboptimal outcomes. 
For instance (as shown in \figureautorefname~\ref{fg:text_weight}), labels like “jackfruit” receive text descriptions related to taste and smell, irrelevant to visual cues, which is less useful in this case. Similarly, the way we prompt images has the same issue, where $p(\cdot)$ is likely to generate unexpected output, such as the areas containing only background information or task-unrelated objects as demonstrated in \figureautorefname~\ref{fg:image_weight}. This motivates us to develop a method to select reliable localized image patches and text descriptions. So this leaves us with another challenge: \emph{how to select reliable ${\bm x}_i$ and ${\bm y}_j$ in the set of localized image patches $p(\bm x)$ and the set of text descriptions $h(\bm y)$, respectively}.

\textbf{Weighted aggregation for cross alignment.} Considering how the semantic relevance between an image and a text is measured using cosine similarity, the question arises: \emph{could this approach also be applied to image-to-image or text-to-text pairs to assess their relevance?} Our empirical studies in Section~\ref{sec:exp} supports the idea. Consequently, we introduce the set of weights for image patches, denoted as $\mathcal{W} = \{w_i\}_{i=1}^{N}$, and for text descriptions, referred to as $\mathcal{V} =\{v_j\}_{j=1}^{M}$. These weights adjust the contribution of each entry in the similar matrix to the similarity score aggregation as follows:
\begin{align}
    &w_i = \frac{\exp\left(s(\bm x, {\bm x}_i |f,g)\right)}{\sum_{l=1}^{N} \exp\left(s(\bm x, {\bm x}_{l} |f,g)\right)},\label{w_i}\\
    &v_j = \frac{\exp\left(s(\bm y, {\bm y}_j |f,g)\right)}{\sum_{l=1}^{M} \exp\left(s(\bm y, {\bm y}_{l} |f,g)\right)}.\label{v_j}
\end{align} 
Then, the score function for the \method is defined as:
\begin{equation}
\label{wca}
    s_{\text{\method}}(\bm x, \bm y |f,g) = \sum_{i=1}^{N} \sum_{j=1}^{M} w_i v_j s(\bm x_i, \bm y_j |f,g).
\end{equation} The underlying idea here is twofold: (i) a high value of $w_i$ indicates that ${\bm x}_i$ is crucial in representing the primary semantic content of image ${\bm x}$, and (ii) a high value of $v_j$ suggests a strong correlation between ${\bm y}_j$ and the label ${\bm y}$.  Essentially, higher weights are indicative of the relative importance of specific image patches or text descriptions within their respective contexts, which is visually demonstrated in Section \ref{sec:exp}. Finally, $s_{\text{\method}}$ can aggregate a more accurate and reliable score. 

\begin{algorithm}[tb]
\footnotesize
   \caption{Weighted Visual-Text Cross Alignment}
   \label{alg:align}
\begin{algorithmic}[1]
   \INPUT A query image $\bm x \in \mathbb{R}^{H\times W \times 3}$; a label set $\mathcal{Y}=\{\bm y^k \}^K_{k=1}$; an LLM model $h(\cdot)$; the number of crops $N$; the number of text prompts $M$; the lower and upper bound $\alpha, \beta$; the crop function $\phi$.
   
   \FOR{$i=1$ to $N$}
   \STATE \textbf{Sample} $\gamma_i \sim U(\alpha,\beta)$
   \STATE \textbf{Let} $n_i = \gamma_i \times \text{min}(W, H)$
   \STATE \textbf{Obtain} $\bm x_i = \phi(\bm x, n_i)$
\STATE \textbf{Compute} $w_i$ according to Eq.~\eqref{w_i}
   \ENDFOR
   \FOR{$k=1$ to $K$}
   \STATE \textbf{Prompt}
   $h(\bm y^k)=\left\{{\bm y}_j^k\right\}_{j=1}^{M}$ 

   \FOR{$j=1$ to $M$}
   \STATE \textbf{Compute} $v_j$ according to Eq.~\eqref{v_j}
   \ENDFOR
   \STATE \textbf{Obtain} $s_\text{WCA}^k$ according to Eq.~\eqref{wca}
   \ENDFOR
   \STATE $k^*=\argmax_{k \in [1..K]} s_\text{WCA}^k$
   \OUTPUT $\bm y^{k^*}$
\end{algorithmic}
\end{algorithm}

\begin{table*}[t]
\centering
\caption{\textbf{Comparison of zero-shot visual classification performance (accuracy in \%) across different image classification benchmarks using three different CLIP models (B/32, B/16, L/14)}. The standard deviation ($\sigma$) of \method’s performance is listed, along with the improvement ($\Delta$) highlighted in green over the top-performing baseline, which is shown as underlined.}
\label{tab:my-table}
\vskip 0.15in
\resizebox{\textwidth}{!}{
\begin{tabular}{lcccccccccccccccccc}
\toprule
 & \multicolumn{3}{c}{ImageNet} & \multicolumn{3}{c}{CUB} & \multicolumn{3}{c}{Oxford Pets} & \multicolumn{3}{c}{DTD} & \multicolumn{3}{c}{Food101} & \multicolumn{3}{c}{Place365} \\ 
 \cmidrule(l){2-4}\cmidrule(l){5-7}\cmidrule(l){8-10}\cmidrule(l){11-13}\cmidrule(l){14-16}\cmidrule(l){17-19} 
\multirow{-2}{*}{Method} & B/32 & B/16 & L/14 & B/32 & B/16 & L/14 & B/32 & B/16 & L/14 & B/32 & B/16 & L/14 & B/32 & B/16 & L/14 & B/32 & B/16 & L/14 \\ \midrule
CLIP & 62.05 & 66.74 & 73.48 & 51.21 & 56.01 & 62.12 & 85.04 & 88.14 & 93.24 & 42.93 & 42.98 & 52.61 & 82.60 & 88.40 & 92.55 & 38.51 & 39.27 & 39.63 \\
CLIP-E & 63.37 & 68.37 & 75.52 & {\ul 52.74} & 56.16 & 62.53 & {\ul 87.38} & 89.10 & 93.62 & 43.83 & 45.27 & 55.43 & 83.93 & 88.83 & 93.07 & 39.28 & 40.30 & 40.55 \\
CLIP-D & 63.01 & 68.04 & 75.03 & 52.69 & {\ul 57.08} & {\ul 63.26} & 84.46 & 87.52 & 93.30 & 44.20 & 46.17 & 55.05 & 84.12 & 88.85 & 93.03 & {\ul 39.90} & 40.34 & 40.55 \\
Waffle & 63.30 & 68.12 & 75.31 & 52.04 & 56.89 & 62.27 & 85.50 & 86.51 & 91.55 & 42.98 & 44.68 & 54.31 & 83.98 & {\ul 89.06} & 93.33 & 39.47 & {\ul 40.76} & {\ul 40.89} \\
CuPL & {\ul 64.37} & {\ul 69.61} & {\ul 76.62} & 49.76 & 56.42 & 62.15 & 87.03 & {\ul 91.14} & {\ul 94.33} & {\ul 47.50} & {\ul 50.53} & {\ul 60.59} & {\ul 84.20} & 88.98 & {\ul 93.37} & 39.08 & 39.83 & 40.77 \\ \midrule
\rowcolor[HTML]{EFEFEF} 
\method & \textbf{66.84} & \textbf{71.08} & \textbf{77.32} & \textbf{56.91} & \textbf{59.78} & \textbf{65.24} & \textbf{89.89} & \textbf{92.23} & \textbf{94.66} & \textbf{49.39} & \textbf{52.79} & \textbf{61.78} & \textbf{86.40} & \textbf{90.01} & \textbf{93.96} & \textbf{40.66} & \textbf{41.43} & \textbf{42.23} \\
$\sigma$ & 0.07 & 0.05 & 0.03 & 0.17 & 0.15 & 0.12 & 0.09 & 0.10 & 0.09 & 0.16 & 0.17 & 0.16 & 0.05 & 0.04 & 0.04 & 0.05 & 0.05 & 0.03 \\
$\Delta$ & {\color[HTML]{036400} \textbf{+2.47}} & {\color[HTML]{036400} \textbf{+1.47}} & {\color[HTML]{036400} \textbf{+0.70}} & {\color[HTML]{036400} \textbf{+4.17}} & {\color[HTML]{036400} \textbf{+2.70}} & {\color[HTML]{036400} \textbf{+1.98}} & {\color[HTML]{036400} \textbf{+2.51}} & {\color[HTML]{036400} \textbf{+1.09}} & {\color[HTML]{036400} \textbf{+0.33}} & {\color[HTML]{036400} \textbf{+1.89}} & {\color[HTML]{036400} \textbf{+2.26}} & {\color[HTML]{036400} \textbf{+1.19}} & {\color[HTML]{036400} \textbf{+2.20}} & {\color[HTML]{036400} \textbf{+0.95}} & {\color[HTML]{036400} \textbf{+0.59}} & {\color[HTML]{036400} \textbf{+0.76}} & {\color[HTML]{036400} \textbf{+0.67}} & {\color[HTML]{036400} \textbf{+1.34}} \\ \bottomrule
\end{tabular}
}
\vskip -0.1in
\end{table*}

\begin{table*}[t]
\centering
\caption{\textbf{Comparison on natural distribution shifts with accuracy (\%) reported}. TP, VP, TTP, and LLM represent textual prompting, visual prompting, test-time promoting, and large language models, respectively. The term ``Tuned" refers to whether the model is fine-tuned on ImageNet. ``Source" refers to in-distribution performance, while ``Target" represents out-of-distribution performance.}
\label{tab:dg-exp}
\vskip 0.15in
\fontsize{7.5pt}{7.5pt}\selectfont
\begin{tabular}{lcccccccc}
\toprule
\multicolumn{1}{c}{} &  &  & Source & \multicolumn{4}{c}{Target} &  \\ \cmidrule(l){4-4}\cmidrule(l){5-8}
\multicolumn{1}{c}{\multirow{-2}{*}{Method}} & \multirow{-2}{*}{Prompts} & \multirow{-2}{*}{Tuned?} & ImageNet & ImageNet-V2 & ImageNet-R & ImageNet-S & ImageNet-A & \multirow{-2}{*}{Average} \\ 
\midrule
CoOp   \cite{zhou2022learning} & TP & \cmark & 71.51 & 64.20 & 75.21 & 47.99 & 49.71 & 61.72 \\
CoCoOp   \cite{zhou2022conditional} & VP+TP & \cmark & 71.02 & 64.07 & 76.18 & 48.75 & 50.63 & 62.13 \\
UPT   \cite{zang2022unified} & VP+TP & \cmark & \textbf{72.63} & 64.35 & 76.24 & 48.66 & 50.66 & 62.51 \\
ProGrad   \cite{zhu2023prompt} & TP & \cmark & {\ul 72.24} & {\ul 64.73} & 74.58 & 47.99 & 49.39 & 61.79 \\
KgCoOp   \cite{yao2023visual} & TP & \cmark & 71.20 & 64.10 & 76.70 & {\ul 48.97} & 50.69 & 62.33 \\ 
\midrule
TPT   \cite{shu2022test} & TTP & \cmark & 69.70 & 64.30 & 73.90 & 46.40 & 53.67 & 61.59 \\
DiffTPT   \cite{feng2023diverse} & TTP & \cmark & 70.30 & \textbf{65.10} & 75.00 & 46.80 & {\ul 55.68} & {\ul 62.58} \\ 
\midrule
CLIP   \cite{radford2021learning} & Hand-crafted & \xmark & 66.74 & 60.83 & 73.96 & 46.15 & 47.77 & 59.09 \\
CLIP-E   \cite{radford2021learning} & Hand-crafted & \xmark & 68.37 & 61.90 & {\ul 77.40} & 47.87 & 49.00 & 60.91 \\
CuPL   \cite{pratt2023does} & LLM-TP & \xmark & 69.61 & 63.27 & 77.10 & 48.80 & 50.77 & 61.91 \\ \midrule
\rowcolor[HTML]{EFEFEF} 
\method (Ours) & LLM-TP+VP & \xmark & 71.08 & 64.71 & \textbf{78.06} & \textbf{50.18} & \textbf{56.13} & \textbf{64.03} \\ \bottomrule
\end{tabular}
\vskip -0.1in
\end{table*}

\textbf{Overall algorithm.}  Algorithm \ref{alg:align} demonstrates how to predict the best match label from $\mathcal{Y}$ for a query image $\bm x$. The algorithm first prompts the image $\bm x$ into multiple localized regions $\{{\bm x}_i\}_{i=1}^{N}$, and assigns a weight $w_i$ to each $\bm x_i$. Similarly, for each label $\bm y^k \in \mathcal{Y}$, $h(\cdot)$ is used to generate $\bm y^k$-related descriptions $\{{\bm y}^k_j\}_{j=1}^{M}$, and $v_j$ is assigned to each $\bm y_j^k$.
The core of the algorithm is the calculation of a cross-alignment score $s_{\text{\method}}^k$. Finally, the algorithm selects the label $\bm y^{k^*}$ that maximizes this cross-alignment score, indicating it as the most suitable label for the image $\bm x$.

\section{Experiments}
\label{sec:exp}

In this section, we evaluate the performance of our method by a series of experiments and various ablation studies.  A detailed insight into our method is also provided. 

\textbf{Datasets.} 
First, we evaluate our method on zero-shot visual classification benchmarks outlined in \citep{menon2022visual}:
\begin{enumerate*}[label=(\roman*)]
  \item ImageNet \cite{deng2009imagenet} for recognizing everyday objects;
  \item CUB for fine-grained classification of birds \cite{wah2011caltech};
  \item Oxford Pets \cite{parkhi2012cats} for common animals;
  \item DTD \cite{cimpoi2014describing} for in-the-wild patterns;
  \item Food101 \cite{bossard2014food} specifically designed for food classification; and
  \item Place365 \cite{zhou2017places} for scene recognition.
\end{enumerate*}

Then we evaluate our method on domain generalization benchmarks in \citep{radford2021learning}, including:
\begin{enumerate*}[label=(\roman*)]
  \item ImageNet-V2 \cite{recht2019imagenet} to evaluate distribution shift from ImageNet;
  \item ImageNet-Sketch \cite{wang2019learning} consisting of black and white sketch images;
  \item ImageNet-A \cite{hendrycks2021natural} for naturally occurring images that are adversarial examples; and
  \item  ImageNet-R \cite{hendrycks2021many} for focusing on art, cartoons, graffiti, and other renditions. 
\end{enumerate*} Each dataset represents a unique distribution shift from ImageNet. This benchmark evaluates the model's robustness in natural distribution shifts.

\textbf{Baselines.} 
In the context of zero-shot visual classification, our evaluation includes a comparison with the following baselines: 
\begin{enumerate*}[label=(\roman*)]
  \item CLIP \cite{radford2021learning}, an approach utilizing a manually created template: ``A photo of \{class\};
  \item An ensemble version of CLIP (CLIP-E) \cite{radford2021learning} employing a variety of manually crafted templates;
  \item  CLIP-D \cite{menon2022visual} leveraging LLMs for the description generation;
  \item CuPL \cite{pratt2023does} known for generating higher quality LLM descriptions in comparison to CLIP-D; and
  \item Waffle \cite{roth2023waffling}, a unique approach that replaces LLM-generated descriptions with randomly generated character and word descriptions.
\end{enumerate*}

Furthermore, we employ the following methods for comparison in domain generalization benchmarks: CoOp \cite{zhou2022learning}, CoCoOp \cite{zhou2022conditional}, UPT \cite{zang2022unified}, ProGrad \cite{zhu2023prompt}, KgCoOp \cite{yao2023visual} and MaPLe \cite{khattak2023maple}. Also, we compare our method to test-time prompting methods, such as TPT \cite{shu2022test} and DiffTPT \cite{feng2023diverse}. Notably, these methods require model fine-tuning, whereas our method operates without any tuning.

\begin{figure}[t]
\begin{center}
\centerline{\includegraphics[width=1.\columnwidth]{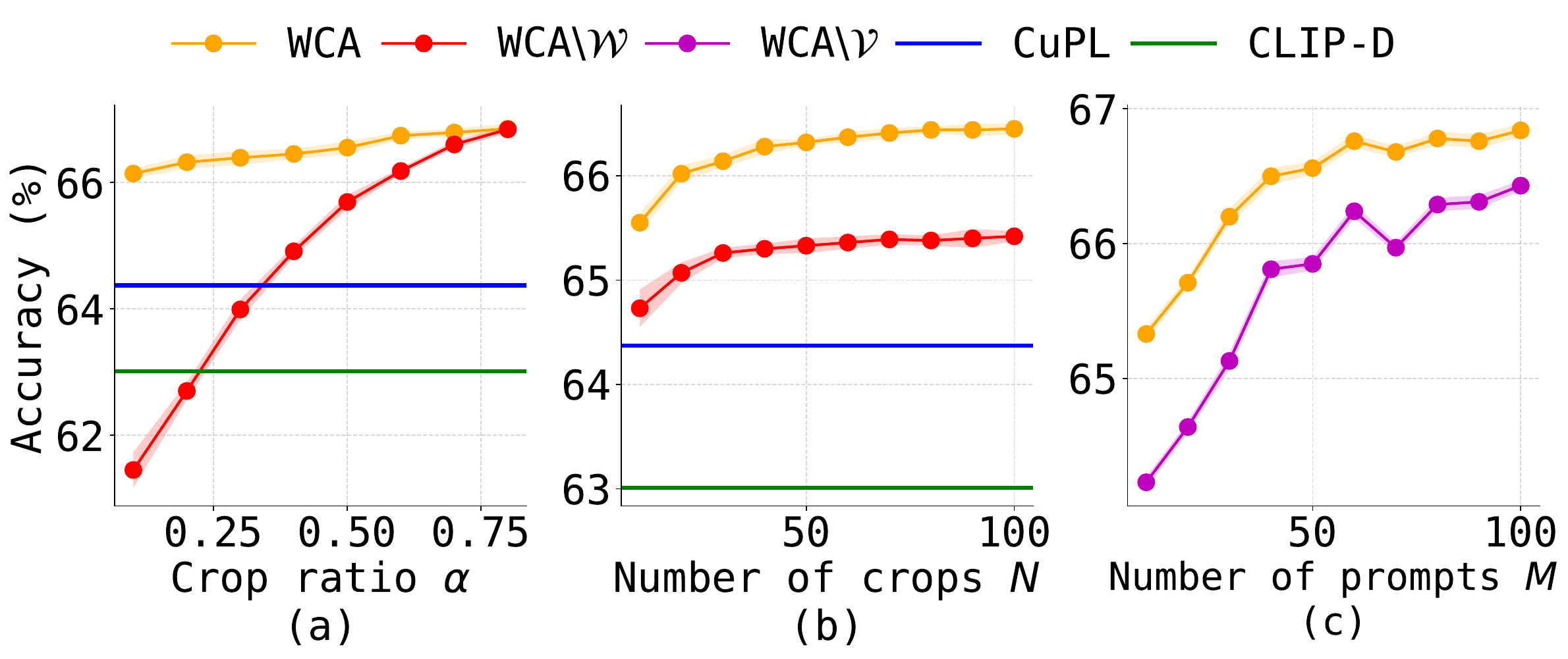}}
\caption{\textbf{Sensitivity analysis of crop ratio $\alpha$, number of crops $N$ and number of prompts $M$}. The shading around curves represents the standard deviation. Note that WCA\textbackslash$\mathcal{W}$ and WCA\textbackslash$\mathcal{V}$ represent the \method without patch weights $\mathcal{W}$ and text weights $\mathcal{V}$, respectively. CuPL \cite{pratt2023does} and CLIP-D \cite{menon2022visual} represent comparative baselines.}
\label{fg:sensitivity}
\end{center}
\vskip -0.3in
\end{figure}

\textbf{Implementation details.} 
We employ a range of VLMs, including CLIP \cite{radford2021learning}, ALIGN \cite{jia2021scaling}, GroupViT \cite{xu2022groupvit} and AltCLIP \cite{chen2022altclip}. Unless specified otherwise, our experiments are conducted using CLIP\footnote{\url{https://github.com/openai/CLIP}} with a backbone of ViT-B/32. All experiments are performed on an NVIDIA A100 GPU. 
Our method incorporates two key parameters: the crop lower and upper bound ($\alpha, \beta$) and the number of crops ($N$). We evaluated various $\beta$ values and observed that larger $\beta$ yields better results as demonstrated in \tableautorefname~\ref{tab:beta} (in Appendix). Thus we set $\beta=0.9$.  In addition, other parameters are generally set to $\alpha = 0.5$, $N = 60$ and $M=50$ across all experiments. 
These values were chosen by our empirical analysis in \figureautorefname~\ref{fg:sensitivity}. 
To optimize computational efficiency, we adopt a strategy where the embedding of an image is pre-computed and stored. This embedding is derived from a weighted average of the embeddings of its localized image patches, a method detailed in Appendices \ref{app:another_view} and \ref{sec:complexity}. This approach guarantees that computational costs do not increase over time, as the embedding is only computed once, which is similar to the technique used in CLIP \cite{radford2021learning} for managing the expenses with prompt ensembling.

 The descriptions used in our study are derived from prior works \citep{menon2022visual, pratt2023does}, which have made progress in automating description generation with minimal human involvement. These works guide LLMs to produce descriptions efficiently by using carefully designed prompts. For example, a prompt from \citep{menon2022visual} asks the model to identify useful features for distinguishing a specific category in a photo. The models then output the visual features associated with the specified category, and these outputs are stored for later use. Additionally, \citet{menon2022visual, pratt2023does} have made certain files containing pre-generated outputs available as open-source resources, serving as the default approach when implementation details are unspecified. Examples using the PaLM model are in \appendixautorefname~\ref{sec:palm}.

\begin{figure*}[t]
\begin{center}
\centerline{\includegraphics[width=\textwidth]{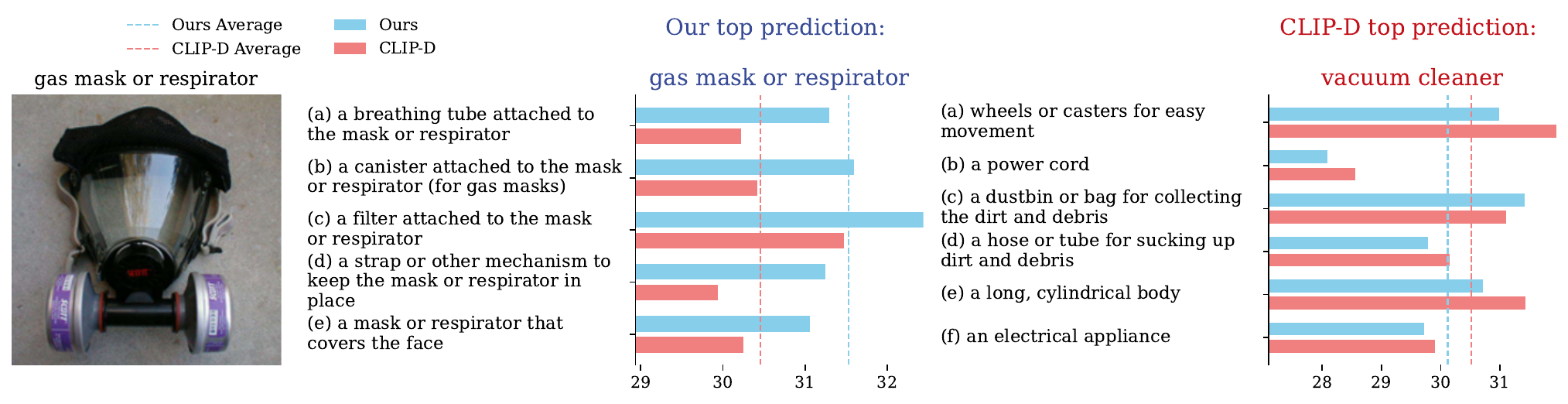}}
\caption{\textbf{We demonstrate the prediction and explanation of our methods and CLIP-D} \cite{menon2022visual}, in identifying and explaining a given image of a gas mask or respirator. The image is analyzed to predict its category, with the scaled cosine similarity scores between the image and various descriptions plotted for each method.}
\label{fg:example}
\end{center}
\vskip -0.3in
\end{figure*}

\textbf{Zero-shot visual classification results.} Table \ref{tab:my-table} showcases the zero-shot visual classification performance comparison across different image classification benchmarks and different model sizes. The results underscore the consistent superiority of \method over established baselines. Furthermore, the results in the table highlight an intriguing trend: the smaller-sized models exhibit more significant performance enhancements compared to their larger counterparts. This phenomenon suggests that while larger models like CLIP are known for their robustness, the relatively smaller models inherently possess more room for improvement, allowing \method to yield substantial gains in accuracy. Moreover, our method excels notably in tasks where CLIP models struggle, indicating that these particularly challenging tasks offer substantial potential for improvement. This observation re-emphasizes the fact that our method excels in addressing complex tasks where there's ample room for advancement, potentially leading to substantial performance gains in domains that pose greater challenges for existing models. Additionally, we show a case where \method makes a correct prediction and its decision is explained by its descriptions in \figureautorefname~\ref{fg:example} and more examples can be found in \appendixautorefname~\ref{app:decisions}.

\textbf{Domain generalization results.} \tableautorefname~\ref{tab:dg-exp} presents a comparison of various methods on their performance across different natural distribution shifts of the ImageNet dataset. Our method surpasses others in in-distribution (Imagenet) performance, except for UPT, and does so without needing fine-tuning data. For out-of-distribution datasets, it excels except on ImageNet-V2, achieving the highest average score, notably on ImageNet-S and ImageNet-A.  This demonstrates that our method is comparable to state-of-the-art in-distribution performance and significantly surpasses them in out-of-distribution scenarios.

\begin{table}[t]
\centering
\caption{\textbf{Ablation study on ImageNet}. Top-1 accuracy (\%) is reported here. The \textbf{bold} value indicates the highest accuracy in each column. The first row serves as the baseline. $\Delta$ shows the mean improvement on top-1 accuracy compared the baseline.}
\label{tab:ablation}
\vskip 0.15in
\fontsize{7.5pt}{7.5pt}\selectfont
\begin{tabular}{cccccccc}
\toprule
\multirow{2}{*}{$p(\cdot)$} & \multirow{2}{*}{$h(\cdot)$} & \multirow{2}{*}{$\mathcal{W}$} & \multirow{2}{*}{$\mathcal{V}$} & \multicolumn{3}{c}{ImageNet} & \multirow{2}{*}{$\Delta$} \\ \cmidrule(lr){5-7}
 &  &  &  & B/32 & B/16 & L/14 &  \\ \midrule
 \multicolumn{4}{c}{Baseline}&   {63.35} & {68.36} & {75.52} & $-$ \\\midrule
 & \cmark &  &  & 64.36 & 69.61 & 76.63 & {\color[HTML]{036400} +1.12} \\
 & \cmark &  & \cmark & 64.77 & 70.09 & 76.68 & {\color[HTML]{036400} +1.44} \\
\cmark &  &  &  & 64.76 & 68.76 & 75.53 & {\color[HTML]{036400} +0.61} \\
\cmark &  & \cmark &  & 65.44 & 69.50 & 76.23 & {\color[HTML]{036400} +1.31} \\
\cmark & \cmark &  &  & 65.51 & 69.72 & 76.34 & {\color[HTML]{036400} +1.45} \\
\midrule
\rowcolor[HTML]{EFEFEF} 
\textbf{\cmark} & \textbf{\cmark} & \textbf{\cmark} & \textbf{\cmark} & \textbf{66.66} & \textbf{71.03} & \textbf{77.33} & \textbf{{\color[HTML]{036400} +2.60}} \\ \bottomrule
\end{tabular}
\vskip -0.2in
\end{table}

\textbf{Ablation study.} The ablation study presented in \tableautorefname~\ref{tab:ablation} systematically shows how various elements impact the top-1 accuracy across three model sizes (B/32, B/16, L/14) for ImageNet classification. The inclusion of each element, either alone or combined, demonstrates differing levels of influence on the model's performance. Notably, when all components are combined, there is a significant improvement in accuracy, indicated by bold values, which show the highest accuracy across all model sizes. 
Specifically, including only $p(\cdot)$ can enhance the performance solely for the B/32 model or potentially worsen it compared to the baseline for B/16 and L/14. This is because the randomness in generating $p(\cdot)$ could still negatively impact the model's performance. However, integrating reweighting parameters $\mathcal{W}$ improves performance consistently by ensuring the selection of only reliable patches.
Moreover, the inclusion of $h(\cdot)$ demonstrates an average improvement of 1.12\% with the support of LLMs, explaining the observed enhancement. In contrast, weighted cropping, involving both $h(\cdot)$ and $\mathcal{W}$, shows a larger improvement of 1.45\% compared to the inclusion of $h(\cdot)$ alone. This emphasizes the efficiency of our method, even without relying on other models or data.

\textbf{Sensitivity analysis.}
 \figureautorefname~\ref{fg:sensitivity}.(a) shows how accuracy correlates with the lower bound parameter $\alpha$. It indicates that increasing $\alpha$ generally leads to a higher accuracy. The line labeled WCA$\setminus\mathcal{W}$ suggests that a small $\alpha$ might result in accuracy falling below the baseline. However, the line WCA demonstrates that $\alpha$ has a minimal impact on performance, suggesting that the factor $\mathcal{W}$ reduces the sensitivity of $\alpha$. \figureautorefname~\ref{fg:sensitivity}.(b) illustrates the effect of the number of image crops $N$ on accuracy. It shows that increasing the number of crops has a positive impact on accuracy. As can be seen, the performance reaches a plateau at $N = 60$. 
 Our observations from \figureautorefname~\ref{fg:sensitivity}.(c) suggest that increasing the number of descriptions generally leads to improved performance, up to a certain threshold. However, when incorporating text description weights $\mathcal{V}$ in WCA, we notice that the performance tends to converge faster compared to the scenario without $\mathcal{V}$. Additionally, for a small number of prompts, not using weights results in lower performance compared to the baseline. This underscores the significance of incorporating text description weights, as they play a crucial role in enhancing the overall performance.

\begin{table*}[t]
\centering
\caption{\textbf{Comparison of time costs between CLIP and WCA methods for different numbers of patches ($N$) in seconds}. The table includes the time for cropping and preprocessing, encoding, and the total time for both methods. It highlights the additional time required by WCA compared to CLIP as the number of patches increases.}
\label{tab:time-cost}
\vskip 0.15in
\fontsize{7.5pt}{7.5pt}\selectfont
\begin{tabular}{@{}lccccccccccc@{}}
\toprule
\multirow{2}{*}{Process Step} & \multirow{2}{*}{CLIP} & \multicolumn{10}{c}{$N$} \\ \cmidrule(l){3-12} 
 &  & 10 & 20 & 30 & 40 & 50 & 60 & 70 & 80 & 90 & 100 \\ \midrule
Crop+Preprocess & 0.0032 & 0.0195 & 0.0394 & 0.0615 & 0.0861 & 0.1096 & 0.1314 & 0.1572 & 0.1811 & 0.2036 & 0.2032 \\
Encoding & 0.0049 & 0.0049 & 0.0050 & 0.0052 & 0.0061 & 0.0068 & 0.0080 & 0.0084 & 0.0083 & 0.0085 & 0.0086 \\\midrule
Total & 0.0081 & 0.0244 & 0.0444 & 0.0666 & 0.0923 & 0.1164 & 0.1394 & 0.1656 & 0.1894 & 0.2121 & 0.2117 \\ \bottomrule
\end{tabular}%
\vskip -0.1in
\end{table*}

\textbf{Revise failure in text descriptions via weighting.} 
Previous studies \cite{menon2022visual} have pointed out several limitations in how LLMs generate descriptions. These models occasionally produce descriptions that are non-visual features. For example, as shown in \figureautorefname~\ref{fg:text_weight}, when GPT-3 \cite{brown2020language} describes a jackfruit, it mentions descriptions associated with taste and smell, which are not part of the visual features. While these descriptions are accurate, they present difficulties for VLMs, which are primarily designed to process and align visual elements. Non-visual descriptions are less useful in this context as they cannot be visually recognized, especially in the scenario where these models come across categories they have never seen before. 

Recent studies \cite{chen2023difference, bielawski2022does, zhang2022visual} have shown that the CLIP model outperforms text-only trained models, such as Bert, in terms of visual understanding. 
CLIP's advanced visual perception enables it to associate text with corresponding visuals in a way that mirrors human perception.
Consequently, we leverage VLMs' visual perception strengths to overcome the shortcomings of LLMs in generating descriptions.
As illustrated in \figureautorefname~\ref{fg:text_weight} (with additional examples in \figureautorefname~\ref{fg:features}), our method successfully identifies these non-visual features and faulty examples, as evidenced by their low similarity scores. This demonstrates the effectiveness of our method in filtering out non-visual or incorrect descriptions, thereby enhancing the accuracy and reliability of the description generation process in alignment with visual content.

\textbf{Explanation of image patch weights.}
Here we explore the efficacy of using image weights. Our chosen technique for visual prompting is random cropping. While this method is straightforward, it inherently carries the risk of randomness, which could negatively impact performance. The goal is to ensure that the random cropping process captures discriminative regions of the image that can then be effectively cross-aligned with textual descriptions.

\begin{figure}[t]
\vskip 0.1in
\begin{center}
\centerline{\includegraphics[width=\columnwidth]{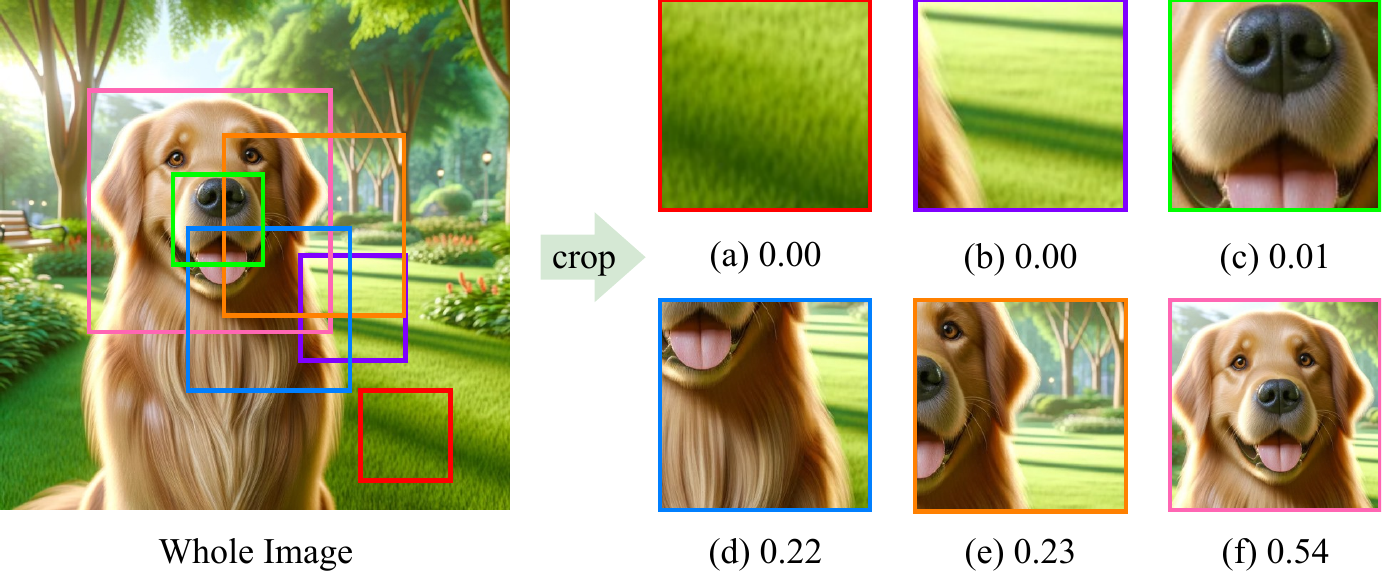}}
\caption{\textbf{Similarity scores between the query image and its localized image patches}. We show that image-patch cosine similarity can filter those patches with less semantic information.}
\label{fg:image_weight}
\end{center}
\vskip -0.3in
\end{figure}

To achieve this, we employ a specific method, as outlined in Eq.~\eqref{w_i}. This equation is designed to evaluate the information degree of different regions within an image. A high score in this context signifies that a particular region of the image is rich in informative content. In \figureautorefname~\ref{fg:image_weight}, we present a series of examples to illustrate this concept. Images (a), (b), and (c) in the figure, for instance, receive scores close to zero. This indicates that the cropped patches from these images contain regions that are not particularly important or informative. In contrast, images (d), (e), and (f) demonstrate a different scenario. These images successfully highlight various characteristic parts of a golden retriever, such as the face, head, and body. The patch weights in these cases are significantly higher, reflecting the discriminative value of these regions. As a result, these images are identified as more informative and thus more suitable for effective cross-alignment with textual descriptions. This approach showcases the potential of image patch weights in enhancing the selection and accuracy of image processing, particularly in aligning images with relevant textual information.

\textbf{Time cost.} 
We break down the time cost of cropping $N$ patches from an image and obtaining their feature embeddings compared to using CLIP. Our experimental dataset consists of $1\,000$ images selected from ImageNet, with the results averaged across these images. As shown in \figureautorefname~\ref{tab:time-cost}, for CLIP, encoding a single image with the CLIP image encoder takes approximately 0.0081 seconds, with 0.0032 seconds for preprocessing and 0.0049 seconds for encoding.
Our method details the time required to crop one image into $N$ patches and encode these patches using the CLIP image encoder through batch processing. The majority of the computational overhead compared to CLIP is attributed to ``Cropping + Preprocessing." In contrast, ``Encoding" time is mitigated by batch processing the patch images.
To optimize computational efficiency and address increased inference time, we pre-compute and store the embeddings of each processed image. Once these embeddings are computed and stored, the inference time is unaffected by ``Cropping + Preprocessing" and ``Encoding." We then only need to perform a dot product between image embeddings and text embeddings, similar to CLIP, which is very fast and takes less than 10 seconds for $50\,000$ images.

\textbf{Additional experiments and analysis}. Additional experiments are detailed in \appendixautorefname~\ref{ap:fur-exp}, including various visual prompting techniques, aggregating strategies, applying \method with various VLMs, and \method with ResNet backbone. Further analysis can be found in \appendixautorefname~\ref{ap:fur-ana}, such as visualization of prompt image embedding. Our discussion on limitations is presented in \appendixautorefname~\ref{app:limitation}.

\section{Conclusion} 
We introduce a method \method, which capitalizes on the precise alignment between localized image areas and finer textual descriptions generated by LLMs, using pre-trained VLMs. By empirically and theoretically demonstrating that finer descriptions align more closely with local image regions, we significantly enhance zero-shot classification performance. Our comprehensive experiments show that \method not only surpasses traditional zero-shot benchmarks but also competes closely with few-shot learning techniques.

\section*{Acknowledgements}
Jinhao Li and Haopeng Li are supported by the Melbourne Research Scholarship.
Feng Liu receives support from the Australian Research Council with grant numbers DP230101540 and DE240101089, and the NSF\&CSIRO Responsible AI program with grant number 2303037.
Sarah Erfani is in part supported by Australian Research Council (ARC) Discovery Early Career Researcher Award (DECRA) DE220100680.

We extend our gratitude to the anonymous reviewers and our colleagues for their insightful comments and suggestions, which greatly improved the quality of this paper. 
Furthermore, we acknowledge the support provided by the University of Melbourne’s Research Computing Services and the Petascale Campus Initiative.

\section*{Impact Statement}
This paper presents work to advance the field of VLMs by enhancing zero-shot performance. Our findings reveal that finer text descriptions align more effectively with localized areas of an image rather than the entire image. This discovery has significant implications for applying VLMs in various real-world scenarios. By improving the zero-shot performance, our method can benefit industries relying on large-scale image data analysis, such as enhancing visual search engines, improving automated image tagging systems, and advancing medical imaging diagnostics.

Societal impacts of our work include democratizing access to powerful AI tools, as our method can deliver high performance even with limited or no labeled data, making advanced VLMs more accessible and usable in resource-constrained environments. Additionally, our approach contributes to the ethical practice of AI by reducing the need for extensive data labeling and minimizing the computational resources required for model training and adaptation, aligning with goals of sustainability and efficiency.

\bibliography{icml2024}
\bibliographystyle{icml2024}

\newpage
\appendix
\onecolumn


\section{Proof of Theorem 1}
\label{proof}

This section outlines the proof of \cref{thm:bigtheorem}.

\begin{definition}
\label{def:1}
(Linear independence \cite{boyd2018introduction}). A collection of vectors $\{\bm a_1,\cdots,\bm a_k\}$ (with $k \geq 1$) is called \emph{linearly independent} if it is not linearly dependent, which means that $\beta_1\bm a_1 + \cdots + \beta_k\bm a_k = \bm 0 \Leftrightarrow \beta_1 = \cdots = \beta_k = 0$.
\end{definition}

\begin{theorem}
\label{thm:cauchy}
(Cauchy–Schwarz inequality \cite{wu2009various}). Let $\bm {u}$  and $\bm {v}$ be arbitrary vectors in an inner product space over the scalar field  $\mathbb{R}$. Then
$ \left|\langle \bm{u} ,\bm{v} \rangle \right| = \|\bm{u} \|\|\bm{v} \| \Leftrightarrow \exists 
\lambda \in \mathbb{R}:\bm{u}=\lambda \cdot \bm{v}.$
\end{theorem}
Let $f: \mathbb{R}^n \rightarrow \mathbb{R}^d$ and $g: \mathbb{R}^m \rightarrow \mathbb{R}^d$ denote two functions. For vectors $\bm{x} \in \mathbb{R}^n, \bm{y} \in \mathbb{R}^m$  with the \emph{same semantic context}, $\cos(f(\bm x), g(\bm y))$ is supposed to be $1$.
\begin{assumption} 
\label{ass:ass1}

Let $f$ be linear functional and satisfies the condition 
\begin{equation}
\label{ass1-2}
    \bm x \neq \bm 0 \Rightarrow f(\bm x) \neq \bm 0.
\end{equation} 
\end{assumption}

\begin{assumption}
Let $\bm x$ can be decomposed into two vectors as follows
\begin{equation}
\label{ass2_1}
    \bm x = \bm x_1 + \bm x_2,
\end{equation}
where 
\begin{align}
&\cos(f(\bm x_1), g(\bm y)) = 1,\label{eq:perfectly}\\
&\cos(f(\bm x_2), g(\bm y)) < 1.
\end{align}
\end{assumption}

Now we begin the proof of \cref{thm:bigtheorem}.

\begin{proof} 
(Contradiction) 
Suppose 
\begin{equation}
\label{eq:support}
  \cos(f(\bm x_1 + \bm x_2), g(\bm y)) = 1,  
\end{equation}
when $\bm x_1$ and $\bm x_2$ satisfy \emph{linear independence} (See Definition \ref{def:1}).

\cref{thm:cauchy} shows that Eq.~\eqref{eq:support} implies $\exists \lambda_1 (\lambda_1 \neq 0)$ such that 
\begin{equation}
\label{eq:imply_1}
    f(\bm x_1 + \bm x_2) = \lambda_1 g(\bm y),
\end{equation}
and Eq.~\eqref{eq:perfectly} implies $\exists \lambda_2 (\lambda_2 \neq 0)$ such that 
\begin{equation}
\label{eq:imply_2}
    f(\bm x_1) = \lambda_2 g(\bm y).
\end{equation}
Combining Eq.~\eqref{eq:imply_1} and Eq.~\eqref{eq:imply_2}, then
\begin{equation}
    f(\bm x_1 + \bm x_2) = \lambda f(\bm x_1),
\end{equation} where $\lambda := \frac{\lambda_1}{\lambda_2}$.
According to $f$ is linear functional, we have
\begin{equation}
    f(\bm x_1) + f(\bm x_2) = \lambda f(\bm x_1), 
\end{equation} which can also be expressed as 
\begin{equation}
\label{eq:1}
    (1-\lambda)f(\bm x_1) + f(\bm x_2) = \bm 0.
\end{equation} 

Then as $f$ is linear functional, we have
\begin{equation}
\label{eq:123}
    f((1-\lambda)\bm x_1 + \bm x_2) = \bm 0.
\end{equation}
Note that the contraposition of the statement in Eq.~\eqref{ass1-2} is 
\begin{equation}
\label{123}
    f(\bm x) = \bm 0 \Rightarrow \bm x = \bm 0.
\end{equation}
Combining Eq. \eqref{eq:123} and Eq. \eqref{123} we have,
\begin{equation}
    (1-\lambda)\bm x_1 + \bm x_2 = \bm 0,
\end{equation} where we have $\beta_1 = 1 - \lambda$ and $\beta_2 = 1$. This violates the condition that $\bm x_1$ and $\bm x_2$ satisfy linear independence. This concludes that the assumption Eq.~\eqref{eq:support} is \emph{false}. 

As the range of $\cos$ is $[-1, 1]$, therefore we have
\begin{equation}
    \cos(f(\bm x_1 + \bm x_2), g(\bm y)) < 1,
\end{equation} then according to Eq.~\eqref{ass2_1} we have 
\begin{equation}
    \cos(f(\bm x), g(\bm y)) < 1,
\end{equation} when $\bm x_1$ and $\bm x_2$ is \emph{linear independent}.
\end{proof}

\section{Further Experiments}
\label{ap:fur-exp}

\subsection{Cross-dataset Generalization Results} 
\label{cross-dataset}

\tableautorefname~\ref{tab:my-table-2} demonstrates the performance of various state-of-the-art fine-tuned methods utilizing 16-shot training data per category and test-time prompting (TPT) methods which require at least one test sample. Notably in terms of in-distribution generalization ability (ImageNet), ours stands out by achieving competitive results without the need for any data-driven parameter tuning. A crucial highlight for the performance in ImageNet is that our method attains comparable results, being just marginally less accurate than CoOp, while we significantly surpass other methods. However, the distinction lies in our approach's superior generalization ability, which is notably stronger than that of other methods. This observation signifies that while our method does not necessitate model fine-tuning, it assures the preservation of the foundation model's generalization capacity. In essence, our method achieves remarkable performance without compromising the model's inherent ability to generalize to new, unseen data within the image distribution.

\subsection{Experiment with Various Visual Prompting Methods}
\label{app:visual}

The \tableautorefname~\ref{tab:visual} is presented to evaluate the top-1 accuracy of image classification in ImageNet under different prompting methods. CLIP serves as the baseline accuracy of 64.37\% without any visual prompting methods. The subsequent columns represent different visual prompting techniques: ``Red Circle", ``Blur", ``Greyscale", and ``Random Crop", which results in a varying impact on the model's performance. Notably, ``Red Circle" and ``Blur" show a slight decrease in accuracy (61.83\% and 61.85\%, respectively) compared to the baseline. ``Greyscale" results in a minor drop in accuracy (62.36\%), indicating that color information might be somewhat relevant to the model's performance. Our method, ``Random Crop" increases the accuracy to 66.84\%, surpassing the baseline and other methods. This indicates that this particular prompting method might be introducing some beneficial variance or focusing the model's attention on more relevant features of the images.

\subsection{Experiment with Various Aggregating Methods}

\begin{table}[t]
\centering
\caption{Comparison of \method with existing approaches on cross-dataset evaluation with fine-tuned methods using 16-shot training data per category (CoOp, CoCoOp, MaPLe) and test-time prompting (TPT, DiffTPT) methods requiring extra parameter turning. We report the top-1 classification accuracy (\%) on each dataset.}
\label{tab:my-table-2}
\vskip 0.15in
\fontsize{8pt}{8pt}\selectfont
\begin{tabular}{lccccccc}
\toprule
\multirow{2}{*}{Method} & \multirow{2}{*}{Tuned?} & Source & \multicolumn{4}{c}{Target} & \multirow{2}{*}{Average} \\ 
\cmidrule(lr){3-3}\cmidrule(lr){4-7}
 &  & ImageNet & DTD & Pets & Food101 & Flowers102 &  \\ \midrule
CoOp \cite{zhou2022learning} & \cmark & \textbf{71.51} & 41.92 & 89.00 & 85.30 & 68.71 & 71.29 \\
CoCoOp \cite{zhou2022conditional} & \cmark & 71.02 & 45.73 & 88.71 & 86.06 & 71.88 & 72.68 \\
MaPLe \cite{khattak2023maple} & \cmark & 70.72 & 46.49 & 90.49 & 86.20 & 72.23 & 73.23 \\
TPT \cite{shu2022test} & \cmark & 69.70 & 46.23 & 86.49 & 86.93 & 69.31 & 71.73 \\
DiffTPT \cite{feng2023diverse} & \cmark & 70.30 & 47.00 & 88.22 & 87.23 & 70.1 & 72.57 \\ \midrule
CLIP-E \cite{radford2021learning} & \xmark & 68.37 & 45.27 & 89.1 & 88.83 & 71.48 & 72.61 \\
CuPL \cite{pratt2023does} & \xmark & 69.91 & 50.53 & 91.14 & 88.98 & 73.39 & 74.79 \\
\rowcolor[HTML]{EFEFEF} 
\method (Ours) & \xmark & 71.08 & \textbf{54.02} & \textbf{91.96} & \textbf{89.98} & \textbf{73.66} & \textbf{76.14} \\ \bottomrule
\end{tabular}
\vskip -0.1in
\end{table}

\begin{table}[ht]
\caption{Top-1 accuracy (\%) of various visual prompting methods in ImageNet. The figure on the left demonstrates the examples of each visual prompting method.}
\vskip 0.15in
    \begin{tabularx}{\textwidth}{@{}l X @{}}
    \includegraphics[width=0.7\textwidth,valign=c]{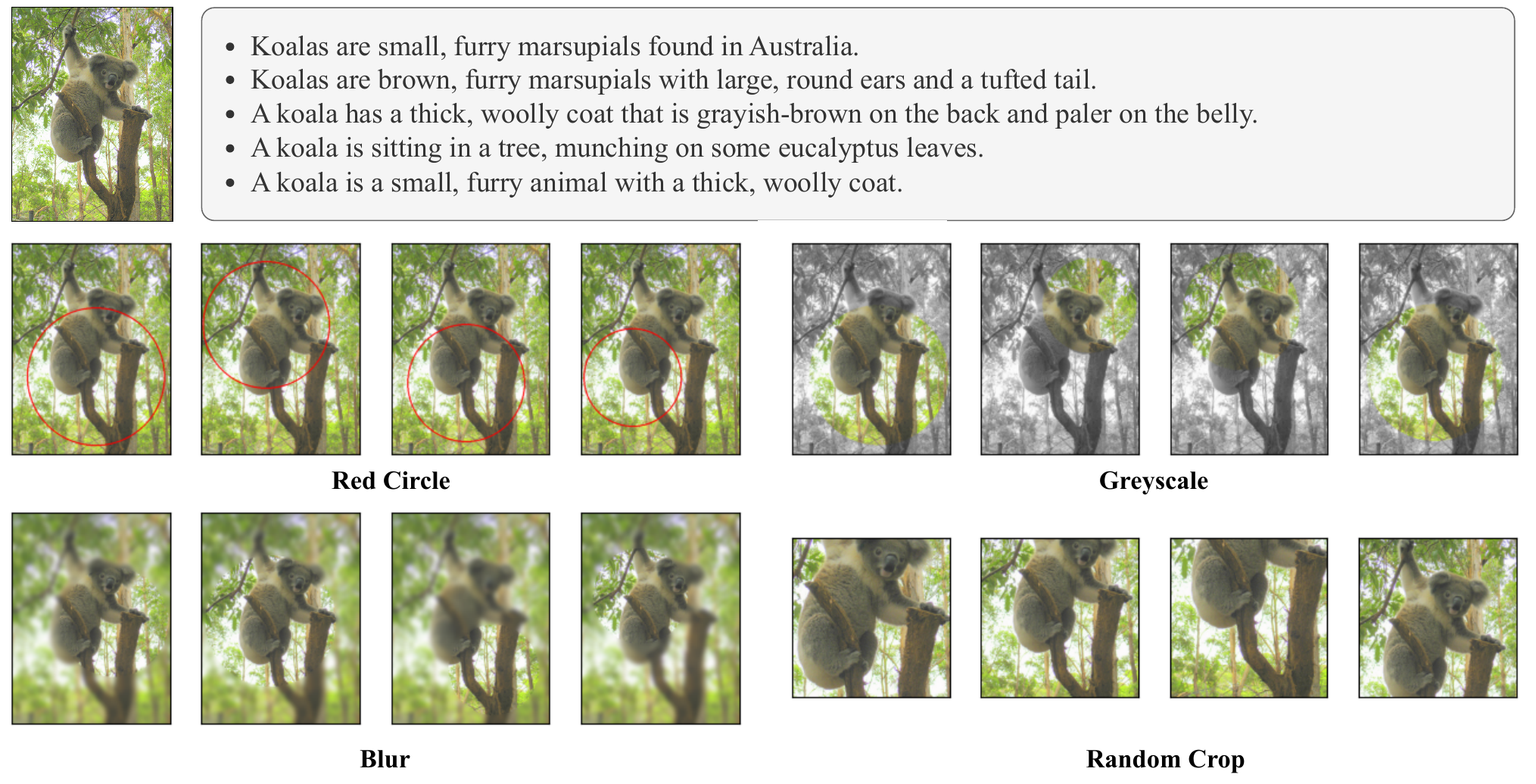}
    &   
        \fontsize{8pt}{8pt}\selectfont
        \begin{tabular}{@{}cc@{}}
        \toprule
        Method & Accuracy \\ \midrule
        Baseline (CLIP) & 64.37 \\
        Red Circle & 61.83 $\pm$ 0.06 \\
        Blur & 61.85 $\pm$ 0.07 \\
        Greyscale & 62.36 $\pm$ 0.07 \\ \midrule
        Random Crop (Ours) & \textbf{66.84} $\pm$ 0.07 \\ \bottomrule
        \end{tabular}
    \end{tabularx}
\label{tab:visual}
\end{table}

\begin{table*}[t]
\centering
\caption{Accuracy (\%) of various aggregating methods on ImageNet with CLIP ViT-B/32, ViT-B/16 and ViT-L/14, where the baseline is selected from the top-performing method from \tableautorefname~\ref{tab:my-table}.}
\vskip 0.15in
\label{tab:aggregation}
\fontsize{8pt}{8pt}\selectfont
\begin{tabular}{@{}ccccc@{}}
\toprule
 \multirow{2}{*}{Backbone}& \multirow{2}{*}{Baseline} & \multicolumn{3}{c}{Aggregation Method} \\ \cmidrule(l){3-5} 
\multirow{-2}{*}{} &  & Max & Mean & \method (Ours) \\ \midrule
ViT-B/32 & 64.37 & 57.53 & {\color[HTML]{000000} 65.51} & \textbf{66.84} \\
ViT-B/16 & 69.61 & 59.94 & 69.72 & \textbf{71.08} \\
ViT-L/14 & 76.62 & 67.99 & 76.34 & \textbf{77.32} \\ \bottomrule
\end{tabular}
\end{table*}

The \tableautorefname~\ref{tab:aggregation} illustrates the zero-shot visual classification performance using various aggregation techniques. It is evident from the table that both Max and Mean aggregation methods are not only less effective compared to our method but also fall behind the baseline in some cases.

\begin{table}[t]
\centering
\caption{Experiment results of various $\beta$ values. $\alpha$ is set to 0.5. This experiment is evaluated in ImageNet for CLIP ViT-B/32.}
\vskip 0.15in
\label{tab:beta}
\fontsize{8pt}{8pt}\selectfont
\begin{tabular}{cccccc}
\toprule
\multirow{2}{*}{$\alpha = 0.5$} & \multicolumn{5}{c}{$\beta$} \\ \cmidrule(l){2-6} 
 & 0.6 & 0.7 & 0.8 & 0.9 & 1 \\ \midrule
Top-1 Accuracy (\%) & 61.79 & 63.2 & 64.45 & 66.84 & 66.06 \\ \bottomrule
\end{tabular}
\end{table}

\subsection{Experiment with Residual Network Backbone}

\begin{table}[t]
\centering
\caption{Zero-shot visual classification top-1 accuracy (\%) of various ResNet backbone CLIP. $\sigma$ represents the standard deviation. $\Delta$ stands for the performance gain achieved by our method over the best baseline, which is denoted as underlined. Each backbone's top score is shown in \textbf{bold}.}
\vskip 0.15in
\label{tab:rn}
\fontsize{8pt}{8pt}\selectfont
\begin{tabular}{lcccccccc}
\toprule
Backbone & CLIP & CLIP-E & CLIP-D & Waffle & CuPL & \method (Ours) & $\sigma$ & $\Delta$ \\ \midrule
ResNet-50 & 58.19 & 59.70 & 59.50 & 60.40 & {\ul 61.32} & \textbf{62.87} & 0.05 &  {\color[HTML]{036400} +1.55} \\
ResNet-101 & 61.22 & 62.33 & 61.92 & 62.97 & {\ul 64.09} & \textbf{64.98} & 0.04 &  {\color[HTML]{036400}+0.89} \\
ResNet-50x4 & 65.52 & 66.54 & 66.09 & 67.11 & {\ul 68.01} & \textbf{68.37} & 0.04 &  {\color[HTML]{036400}+0.36} \\ \bottomrule
\end{tabular}
\vskip -0.1in
\end{table}

We further conduct the experiments with the Residual Network (ResNet) backbone in ImageNet as shown in \tableautorefname~\ref{tab:rn}. We can see our method consistently outperforms other baselines.

\subsection{Exploring Different VLM Architectures}

\begin{table}[t]
\centering
\caption{Top-1 accuracy (\%) for zero-shot visual classification conducted on various VLMs in ImageNet. Each column represents a different prompting method, e.g., CLIP refers to ``A photo of \{label\}". $\sigma$ represents the standard deviation. $\Delta$ stands for the performance gain achieved by our method over the best baseline, which is denoted as underlined. Each VLM's top score is shown in \textbf{bold}.}
\vskip 0.15in
\label{tab:vlm}
\fontsize{8pt}{8pt}\selectfont
\begin{tabular}{lcccccccc}
\toprule
VLM & CLIP & CLIP-E & CLIP-D & Waffle & CuPL & \method (Ours) & $\sigma$ & $\Delta$ \\ \midrule
ALIGN   \cite{jia2021scaling} & 65.24 & 65.79 & 65.08 & 65.22 & 66.24 & \textbf{66.77} & 0.09 & {\color[HTML]{036400} +0.53} \\
AltCLIP   \cite{chen2022altclip} & 73.79 & 74.86 & 74.48 & 74.29 & 75.74 & \textbf{76.20} & 0.04 & {\color[HTML]{036400} +0.46} \\
GroupViT   \cite{xu2022groupvit} & 37.11 & 42.72 & 40.10 & 42.42 & 44.53 & \textbf{45.27} & 0.05 & {\color[HTML]{036400} +0.74} \\ \bottomrule
\end{tabular}
\vskip -0.1in
\end{table}

This section is dedicated to an ablation study involving various VLMs characterized by distinct model architectures and pre-training datasets. Specifically, we examine models such as ALIGN \cite{jia2021scaling}, AltCLIP \cite{chen2022altclip}, and GroupVit \cite{xu2022groupvit}. The outcomes of this exploration are detailed in \tableautorefname~\ref{tab:vlm}. A critical observation from this study is that our method consistently outperforms others even under different VLMs. This consistency underscores the adaptability and effectiveness of our methodology when applied to a diverse range of VLM architectures.

\subsection{Incorporating the Entire Image Features.} 
We have explored the possibility of incorporating the entire image into our method. We devised a new scoring function for the form:
$\lambda \cdot \text{Sim}(x,h(y)) + (1-\lambda) \cdot \text{Sim}(p(x),h(y)),$
where $x$ represents the entire image, $p(x)$ denotes the patch images, $h(y)$ signifies the LLM-generated text descriptions and $\lambda$ serves as a hyperparameter controlling the balance between scores related to the entire image, and patches. However, after thorough experimentation as shown in the table below, we found that the performance improvement achieved by this modification was marginal (0.06\% when $\lambda=0.1$). Moreover, integrating the entire image into the methodology introduced an additional parameter ($\lambda$) and complexities without significant enhancement in performance. 

\begin{table}[t]
\centering
\caption{Impact of varying $\lambda$ on accuracy (\%). The table shows the accuracy values for different $\lambda$ settings, indicating how the accuracy decreases as $\lambda$ increases from 0.0 to 1.0.}
\vskip 0.15in
\label{tab:lambda}
\fontsize{8pt}{8pt}\selectfont
\begin{tabular}{@{}lccccccccccc@{}}
\toprule
$\lambda$ & 0.0 & 0.1 & 0.2 & 0.3 & 0.4 & 0.5 & 0.6 & 0.7 & 0.8 & 0.9 & 1.0 \\ \midrule
Accuracy & 66.84 & 66.90 & 66.88 & 66.77 & 66.65 & 66.48 & 66.28 & 65.99 & 65.65 & 65.28 & 64.83 \\ \bottomrule
\end{tabular}%
\vskip -0.1in
\end{table}

\subsection{Experiment with Google PaLM}

\begin{table}[t]
\centering
\caption{Experiment results comparing the accuracy (\%) of different methods using Google LLM PaLM. The table shows the accuracy percentages for CLIP, CuPL-PaLM, and WCA-PaLM.}
\label{tab:cupl-w}
\vskip 0.15in
\fontsize{8pt}{8pt}\selectfont
\begin{tabular}{cccc}
\toprule
Method & CLIP & CuPL-PaLM & \method-PaLM \\ \midrule
Accuracy (\%) & 62.05  & 62.16 & \textbf{64.03}\\ \bottomrule
\end{tabular}%
\vskip -0.1in
\end{table}

Despite observing a performance improvement using GPT-3 \cite{brown2020language}, we further explored the capabilities of other large-scale language model experiments, such as Google PaLM API \cite{anil2023palm}. We fetched the descriptions for ImageNet classes with the same prompt used in CuPL \cite{pratt2023does}. The outcome shows that the PaLM-based CLIP model achieved a 62.16\% accuracy, marginally exceeding the original CLIP's performance by 0.05\%. By incorporating description weighting as Eq.~\ref{v_j}, we managed to improve the performance to 64.03\%. This improvement highlights the effectiveness of the weighted approach regarding the situation for some reasons LLMs may provide less useful descriptions. 

\begin{table}[t]
\centering
\caption{Computation analysis. This table outlines CLIP, 16-shot CoOp, and our method, which demonstrates that our method is very efficient compared to the prompting turning method, and achieves comparable results. Even though zero-shot CLIP is the most efficient but underperforms other methods. }
\label{tab:time}
\vskip 0.15in
\fontsize{8pt}{8pt}\selectfont
\begin{tabular}{lccc}
\toprule
Method & \#Params & Accuracy (\%) & Time (hh:mm:ss) \\ \midrule
CLIP-D & 0 & 63.01 & 00:00:35 \\
CoOp & 2,048 & 66.85 & 11:34:05 \\ \midrule
\rowcolor[HTML]{EFEFEF} 
\method (Ours) & 0 & 66.84 & 00:06:42 \\ \bottomrule
\end{tabular}
\end{table}

\section{Further Analysis and Discussion}
\label{ap:fur-ana}

\begin{figure*}[t]
\vskip 0.2in
\begin{center}
\centerline{\includegraphics[width=0.9\textwidth]{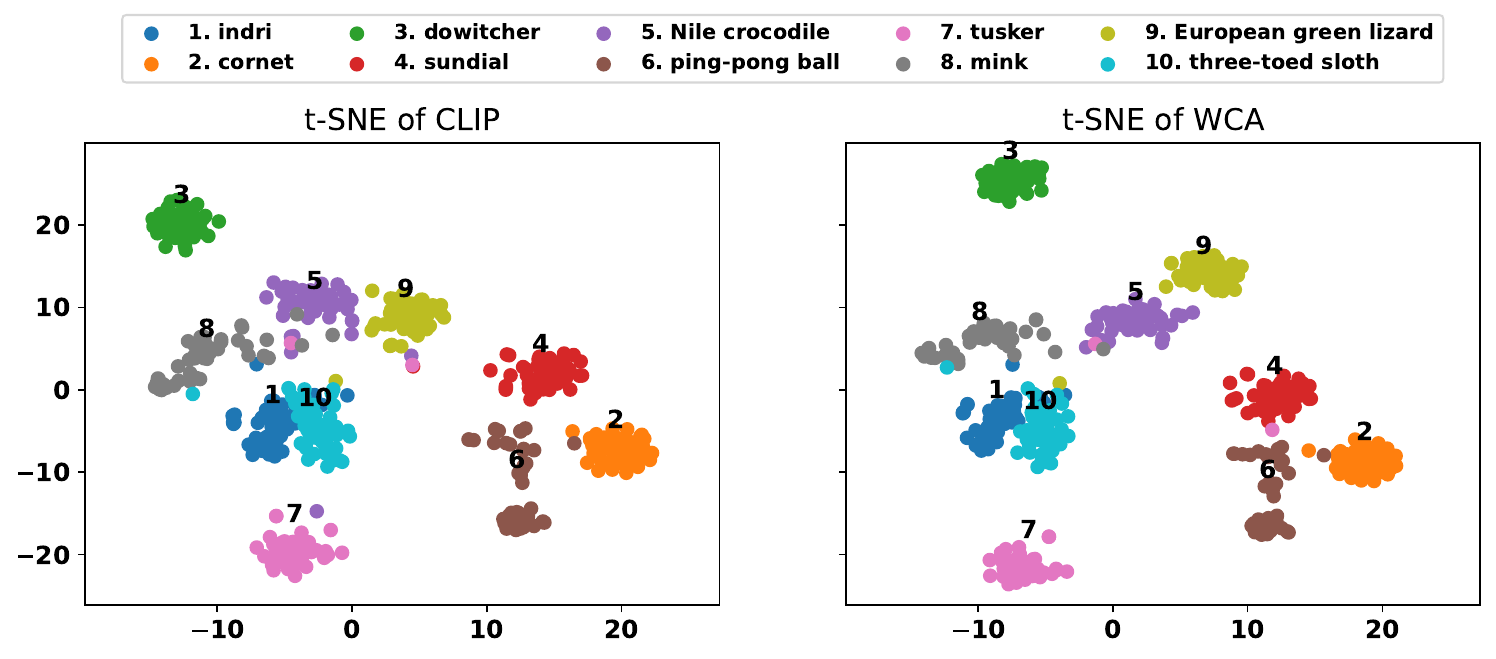}}
\caption{Visualization of t-SNE plots comparing CLIP and \method image embeddings of 10 classes in ImageNet. Each plot represents 50 samples per class, showing the spatial distribution of the embeddings. The classes include indri, cornet, dowitcher, sundial, Nile crocodile, ping-pong ball, tusker, mink, European green lizard, and three-toed sloth. The left plot illustrates the t-SNE of CLIP embeddings, while the right plot shows the t-SNE of WCA embeddings, highlighting the differences in how each model represents the image data.}
\label{fg:tnse}
\end{center}
\vskip -0.4in
\end{figure*}

\begin{figure*}[t]
\vskip 0.2in
\begin{center}
\centerline{\includegraphics[width=0.9\textwidth]{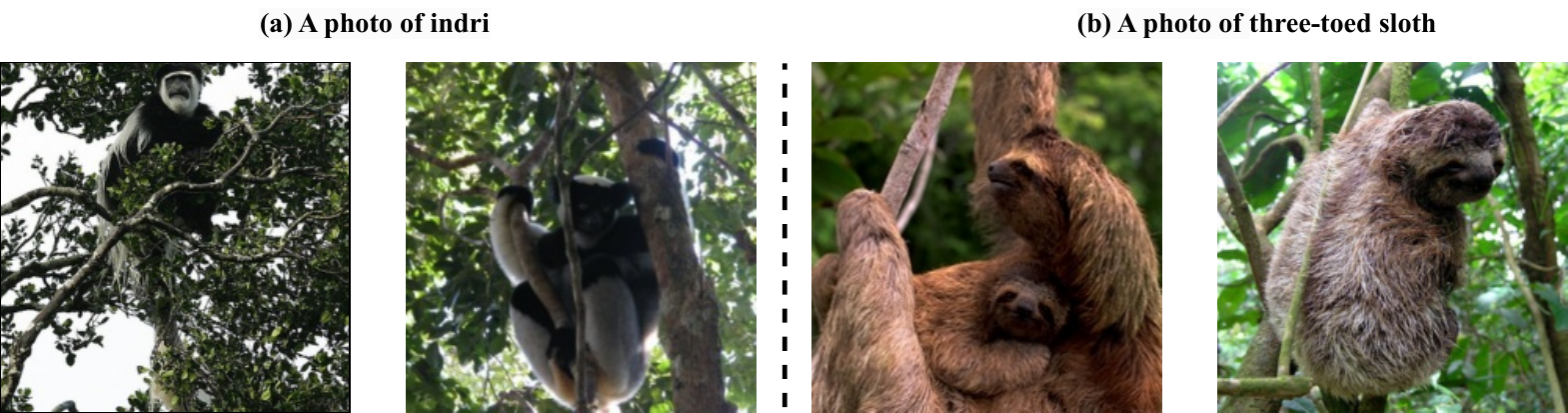}}
\caption{Example images of two animal species from ImageNet: (a) indri, shown in two images capturing its arboreal lifestyle, typically found in the forest canopies of Madagascar; (b) three-toed sloth, depicted in its natural habitat, displaying its characteristic slow movement and hanging posture in the trees of Central and South American rainforests. These images illustrate the visual diversity and distinct ecological niches of each species.}
\label{fg:indr}
\end{center}
\vskip -0.2in
\end{figure*}

\begin{figure*}[t]
\begin{center}
\centerline{\includegraphics[width=0.9\textwidth]{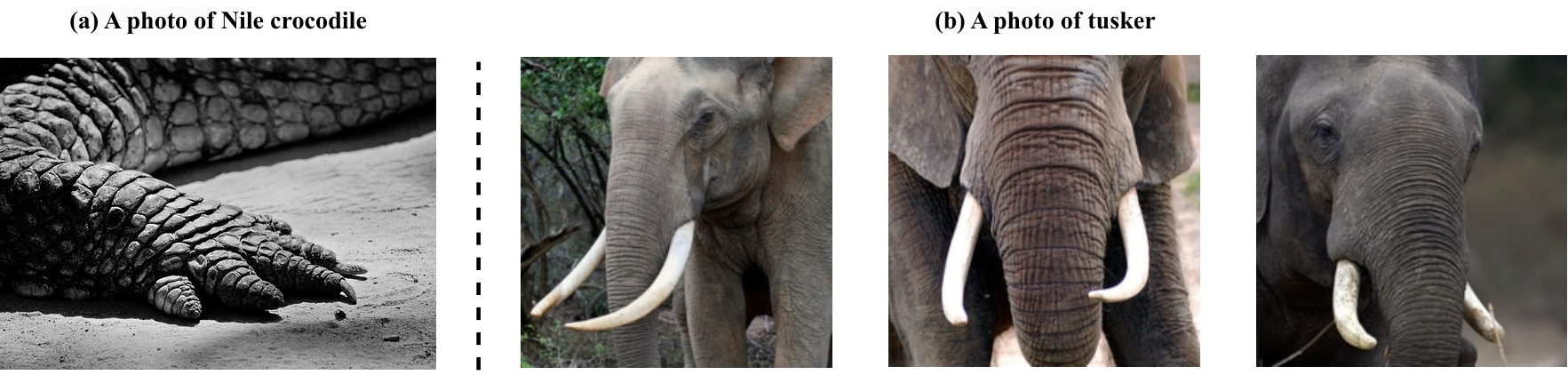}}
\caption{Example images of two animal species from ImageNet: (a) Nile crocodile, depicted with a focus on its textured scales and powerful limbs, highlighting its adaptations for an aquatic lifestyle; (b) tusker, shown in three images featuring close-ups of its large tusks and distinctive features, emphasizing its status as a majestic and significant member of the elephant family. These images illustrate the visual details and distinguishing characteristics of each species.}
\label{fg:tusker}
\end{center}
\vskip -0.4in
\end{figure*}
 
\subsection{Visualisation of Prompt Image Embedding}
\label{app:another_view}

\method is defined as in Eq.~\eqref{wca}, and here we look at another view of our method:
\begin{align}
 s_{\text{\method}}(\bm x, \bm y |f,g) &= 
 \sum_{i=1}^{N} \sum_{j=1}^{M} w_i v_j \frac{f(\bm x_i)^{\mathrm{T}} g(\bm y_j)}{\Vert f(\bm x_i)\Vert \Vert g(\bm y_j)\Vert}\notag\\
 &=\left(\sum_{i=1}^{N} w_i \frac{f(\bm x_i)}{\Vert f(\bm x_i)\Vert }\right)^{\mathrm{T}}\left(\sum_{j=1}^{M} v_j \frac{g(\bm y_j)}{\Vert g(\bm y_j)\Vert}\right)\notag\\
&= \underbrace{\left(\sum_{i=1}^{N}  p_if(\bm x_i)\right)^{\mathrm{T}}}_{\text{Augmented Image Embedding}} \underbrace{\left(\sum_{j=1}^{M}  q_jg(\bm y_j)\right)}_{\text{Augmented Text Embedding}}= \bm k^{\mathrm{T}} \bm t \notag \label{k}
\end{align}
where $p_i:=\frac{w_i}{\Vert f(\bm x_i)\Vert },q_j:=\frac{v_j}{\Vert g(\bm y_j)\Vert}\in \mathbb{R}$, and $\bm k:=\sum_{i=1}^{N}  p_if(\bm x_i), \bm t:=\sum_{j=1}^{M}  q_jg(\bm y_j)\in \mathbb{R}^d$. Based on the derivation, our visual-text alignment score is equivalent to the inner product of the augmented visual embedding $\bm k$ and text embedding $\bm t$. The augmented embeddings are computed as the weighted sum of the embeddings of the image patches/descriptions. $\bm k$ and $\bm t$ can be pre-computed to be stored for quick access in later use.

We employ t-SNE \cite{van2008visualizing} to compare the image embedding of CLIP and our method, using a dataset comprising 500 samples from 10 selected ImageNet classes. As illustrated in \figureautorefname~\ref{fg:tnse}, our method demonstrates more distinct class boundaries, particularly for classes 3 (dowitcher) and 7 (tusker), and a clear separation between classes 1 (indri) and 10 (three-toed sloth). Focusing on classes 1 and 10, which are visually similar as both species climb trees, we observe in \figureautorefname~\ref{fg:indr} that CLIP tends to misclassify an indri as a sloth due to its emphasis on general semantic information. Our method, however, utilizes localized visual prompting to effectively discount tree-related features, enhancing accuracy. Additionally, we explore an outlier purple point near class 7, shown in \figureautorefname~\ref{fg:tusker}, which, despite being an anomaly, aligns closely with the Tusker group images.

\begin{figure*}[t]
\vskip 0.2in
\begin{center}
\centerline{\includegraphics[width=0.9\textwidth]{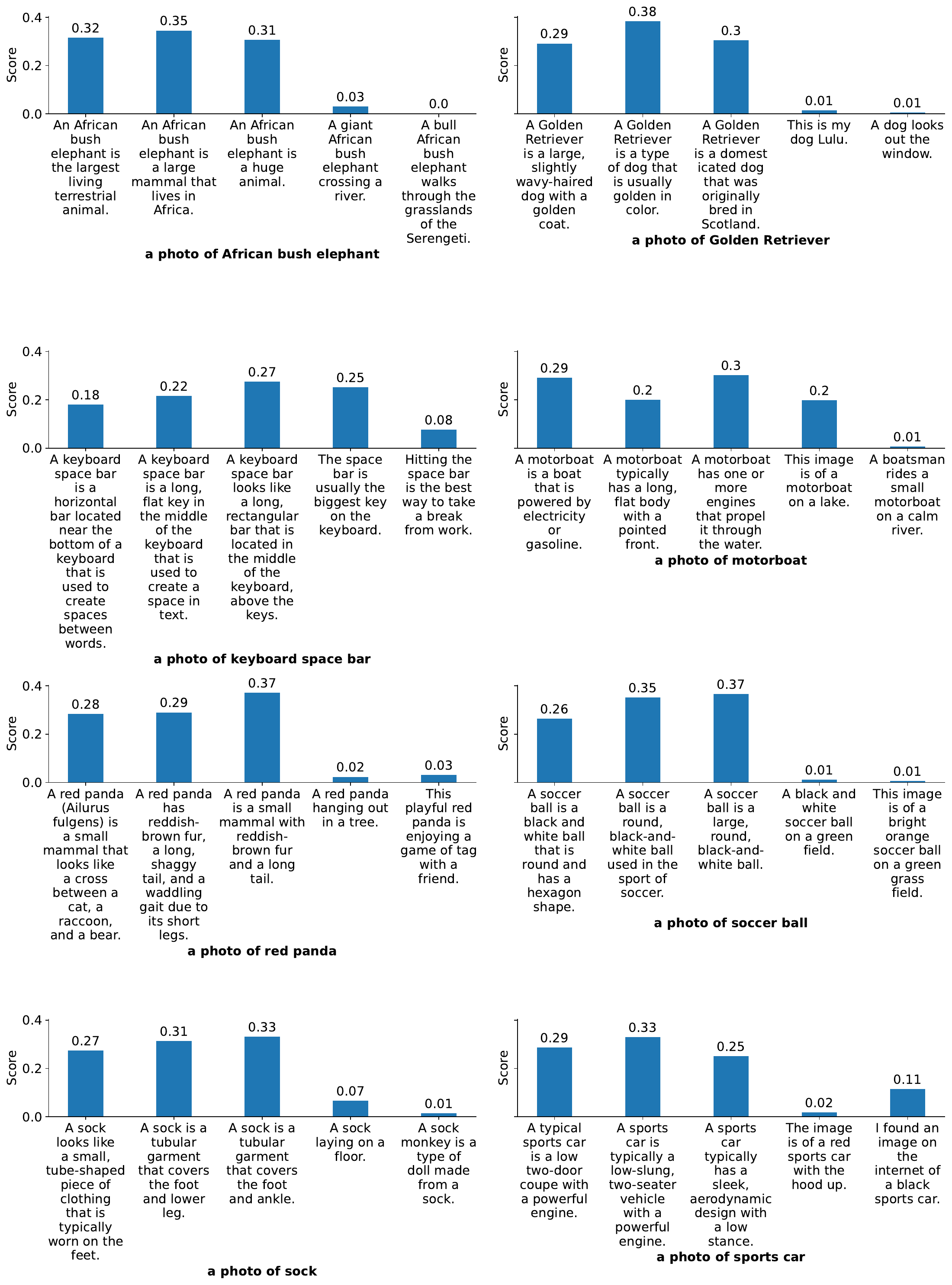}}
\caption{Feature plots comparing the textual alignment scores for different categories: African bush elephant, Golden Retriever, keyboard space bar, motorboat, red panda, soccer ball, sock, and sports car. Each plot shows the alignment scores for five different textual descriptions, highlighting the variation in performance across different categories.}
\label{fg:features}
\end{center}
\vskip -0.4in
\end{figure*}

\begin{figure*}[t]
\vskip 0.2in
\begin{center}
\centerline{\includegraphics[width=0.9\textwidth]{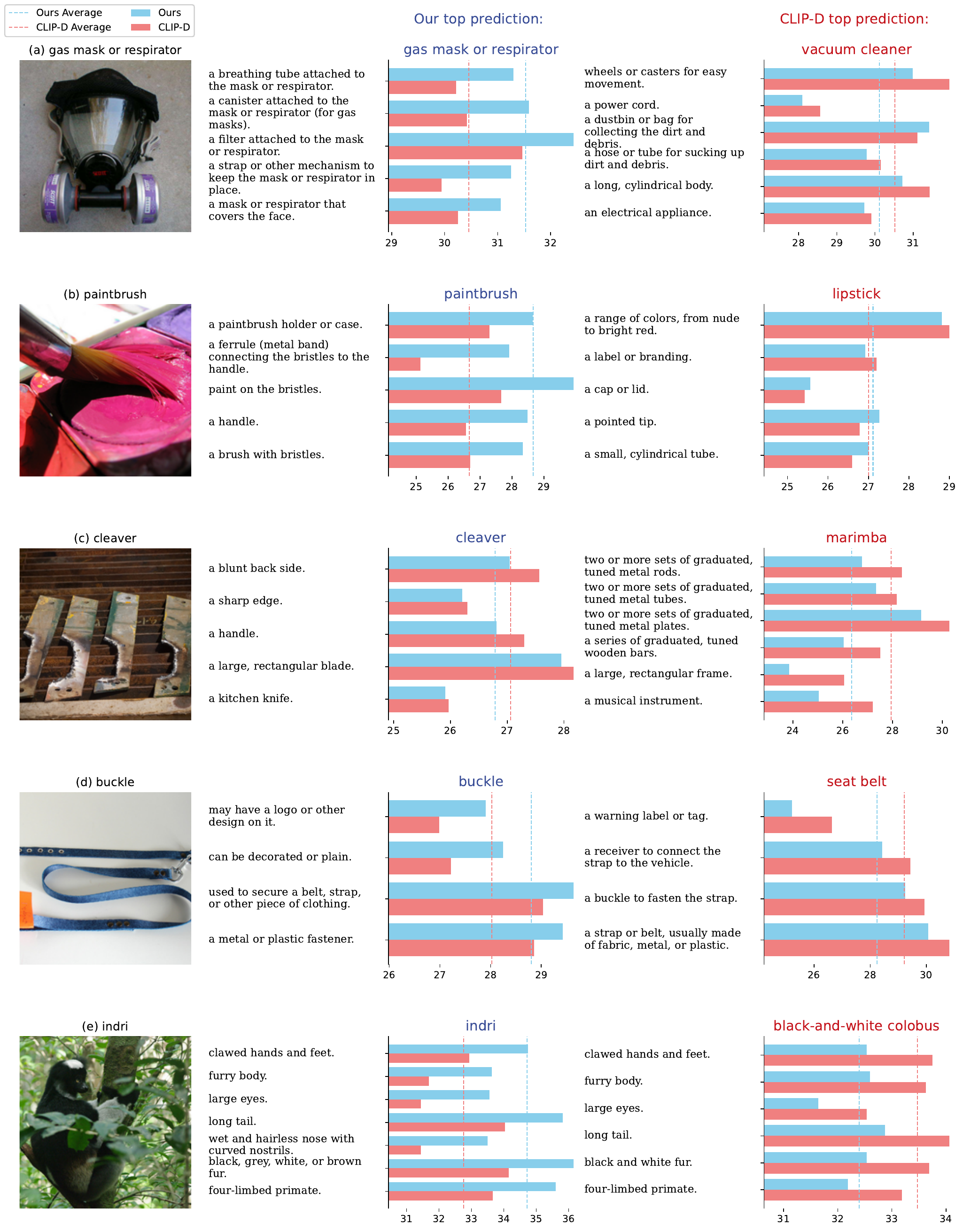}}
\vskip -0.1in
\caption{An illustration of the prediction and explanation of comparison between CLIP-D and our methods. The value in the plot represents the similarity score (higher denotes a high similarity). We can use the LLM-generated descriptions to explain the decisions by the model. For example, the top row means our method predicts it correctly as a gas mask as our method can recognize a filter in the image, while CLIP-D recognizes it as a vacuum cleaner.}
\label{fg:explain_1}
\end{center}
\vskip -0.2in
\end{figure*}

\subsection{Complexity Analysis.}
\label{sec:complexity}

 The computational analysis in \tableautorefname~\ref{tab:time} offers insights into the efficiency of our method. While CLIP stands out for its computational efficiency, it falls short in performance compared to other methods. In contrast, CoOp with its high computational cost only achieves similar performance levels to \method. Our method, \method, strikes a commendable balance between computational load and performance efficiency. It surpasses CLIP in terms of performance while maintaining a significantly lower computational complexity compared to CoOp. This analysis underscores \method's effectiveness as a more optimized solution in the vision-language modeling space, offering a desirable compromise between computational demands and model performance. 

 This observation highlights a key strategy for optimizing the time complexity of the \method method. The primary source of computational demand stems from the image-prompting process. However, this challenge can be effectively mitigated by pre-computing and storing the embeddings of image prompts on a hard drive. By implementing this approach, the computation cost of \method can be brought in line with that of CLIP, effectively neutralizing the additional computational overhead associated with our method. This not only enhances the efficiency of \method but also preserves its superior performance capabilities, making it a highly practical and competitive option in the realm of vision-language models.

\section{Limitation}
\label{app:limitation}

\begin{wrapfigure}{r}{0.5\textwidth} 
    \vskip -0.1in
    \centering
    \includegraphics[width=0.5\textwidth]{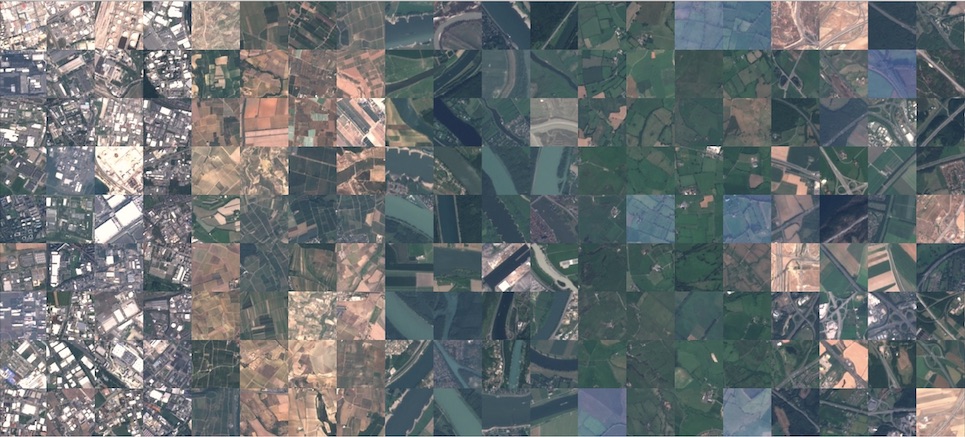}
    \vskip -0.1in
    \caption{Overview of the EuroSAT dataset.}
\label{fg:eurosat}
\end{wrapfigure} Our approach, while effective in certain scenarios, particularly in object recognition tasks such as identifying an image containing a dog, exhibits limited success in contexts requiring comprehensive image understanding. An example of this can be seen with the EuroSAT dataset \cite{helber2018introducing}, which is used for land use and land cover classification. This dataset demands a holistic grasp of the entire image, a requirement evident in \figureautorefname~\ref{fg:eurosat}. The performance in this context challenges our initial assumption that an area of the image would align perfectly with its textual description as described in \cref{thm:bigtheorem}. It suggests that further refinements or a different approach might be necessary for tasks that require a deeper, more holistic understanding of images, as opposed to those that focus on identifying individual objects.

We have also identified another limitation, particularly when dealing with images containing multiple objects of varying sizes. For example, if the task is to identify a small cat in an image crowded with larger dogs, the patch weights might inadvertently emphasize the dogs while downplaying the cat, potentially hindering performance. Additionally, the current approach for text weighting, which relies on cosine similarity to a base description, such as ``a photo of a \{category\}", might not always be optimal, resulting in sub-optimal text weighting.

\begin{figure*}[t]
\begin{center}
\centerline{\includegraphics[width=0.9\textwidth]{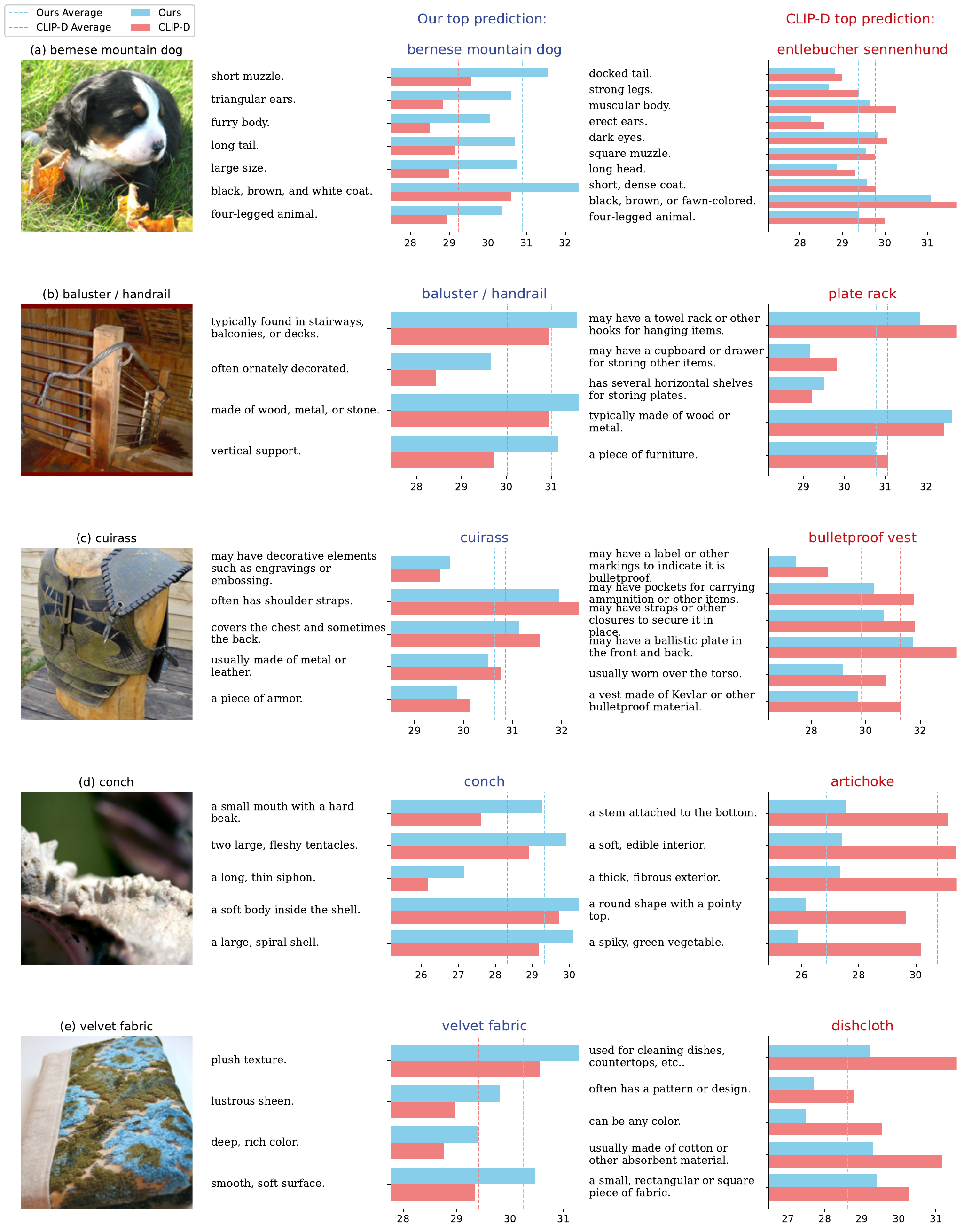}}
\vskip -0.1in
\caption{This figure is set in the same context as \figureautorefname~\ref{fg:explain_1} but with different images.}
\label{fg:explain_2}
\end{center}
\vskip -0.3in
\end{figure*}

\section{Example PaLM Generated Prompts}
\label{sec:palm}

To generate these prompts, we utilized the PaLM model \cite{anil2023palm} with a specific configuration aimed at producing diverse and detailed responses. Here is a step-by-step breakdown of the process:
\begin{enumerate}
\item \textbf{Template Selection:} We started by selecting appropriate templates for generating the prompts. These templates were based on the structure provided by \cite{pratt2023does}, ensuring consistency and clarity in the generated prompts.
\item \textbf{Prompt Generation:} For each category, such as the ``Koala'' category, we fed the selected templates into the model. The prompt question, in this case, was ``Describe what a(n) \{koala\} looks like.'' The model then generated multiple responses based on this prompt.
\item \textbf{Response Collection:} The model was set to generate around eight responses for each prompt. These responses were collected and reviewed to ensure they met the criteria of being descriptive and relevant.
\item \textbf{Post-processing:} After generating the responses, minor post-processing was performed to format the output for clarity and presentation. This included organizing the responses into a list format and ensuring they adhered to the prompt’s context.
\end{enumerate}
Below is an example of the generated descriptions for each prompt:

\begin{tcolorbox}[left=2pt,right=2pt,top=2pt,bottom=2pt]
\texttt{\textbf{Prompt: Describe what a(n) {koala} looks like.}}

\vspace{0.3em} 

\texttt{\textbf{Responses:}}
\begin{enumerate}[label=\texttt{\arabic*.},itemsep=-0.3em]
\item\texttt{Koalas are small, furry marsupials found in Australia.}
\item\texttt{A koala is a small, tree-dwelling marsupial found in Australia.}
\item\texttt{A koala looks like a small, stocky bear with a large head and a long tail.}
\item\texttt{Koalas are small, furry marsupials that are found in Australia.}
\item\texttt{A koala is a small, furry animal that lives in Australia.}
\item\texttt{A koala is a small, furry animal with a large head and a long tail.}
\item\texttt{A koala is a small, furry animal with a thick, woolly coat.}
\item\texttt{Koalas are small, furry marsupials found in Australia.}
\end{enumerate}
\end{tcolorbox}

\section{Examples of Decisions and Justifications}
\label{app:decisions}

As \citet{menon2022visual} discussed, LLM-based CLIP not only provides better performance but also explains the model. Therefore, we randomly selected 10 examples in ImageNet, where we made correct predictions but not for our baseline CLIP-D as shown in \figureautorefname~\ref{fg:explain_1} and \ref{fg:explain_2}. This illustrates the effectiveness of our method. For instance, the top row in  \ref{fg:explain_1} describes a photo of a gas mask, where CLIP-D incorrectly predicts that as a vacuum cleaner, CLIP-D witnesses a high score for its description ``wheels or casters for easy movement.", while our model correctly predicts it as gas mask since our model shows the high scores for the descriptions of gas mask, such as ``a filter attached to the mask". This means our method can recognize this feature inside this photo.



\end{document}